\documentclass{article}

\usepackage{arxiv}

\usepackage[utf8]{inputenc} 
\usepackage[T1]{fontenc}    
\usepackage{hyperref}       
\usepackage{url}            
\usepackage{booktabs}       
\usepackage{amsfonts}       
\usepackage{nicefrac}       
\usepackage{microtype}      
\usepackage{lipsum}		
\usepackage{graphicx}
\usepackage{natbib}
\usepackage{doi}
\usepackage{amsmath}
\usepackage{mathtools}
\usepackage[ruled,vlined]{algorithm2e}
\SetKw{KwBy}{by}
\SetKwInput{KwInput}{Input}
\SetKwInput{KwOutput}{Output}
\usepackage[flushleft]{threeparttable}
\usepackage{longtable}
\usepackage{diagbox}
\usepackage{chngpage}
\newcommand*{\thead}[1]{%
\multicolumn{1}{c}{\bfseries\begin{tabular}{@{}c@{}}#1\end{tabular}}}
\usepackage{pdflscape}
\usepackage{float}
\usepackage{multirow}
\usepackage{subcaption}

\title{Individualized Risk Assessment of Preoperative Opioid Use by Interpretable Neural Network Regression}

\author{Yuming Sun\\
	Department of Biostatistics\\
	University of Michigan, Ann Arbor\\
	\texttt{yumsun@umich.edu} \\
	\And
	Jian Kang\\
    Department of Biostatistics\\
	University of Michigan, Ann Arbor\\
	\texttt{jiankang@umich.edu}
	\And
	Chad Brummett\\
	Department of Anesthesiology\\
	University of Michigan, Ann Arbor\\
	\texttt{cbrummet@med.umich.edu}
	\And
	Yi Li\\
	Department of Biostatistics\\
	University of Michigan, Ann Arbor\\
	\texttt{yili@umich.edu}
}

\date{}

\hypersetup{
pdftitle={A template for the arxiv style},
pdfsubject={q-bio.NC, q-bio.QM},
pdfauthor={David S.~Hippocampus, Elias D.~Striatum},
pdfkeywords={First keyword, Second keyword, More},
}

\begin{document}
\maketitle

\begin{abstract}
	Preoperative opioid use has been reported to be associated with higher preoperative opioid demand, worse postoperative outcomes, and increased postoperative healthcare utilization and expenditures. Understanding the risk of preoperative opioid use helps establish patient-centered pain management. In the field of machine learning, deep neural network (DNN)  has emerged as a powerful
means for risk assessment because of its superb prediction power; however, the blackbox algorithms may  make the results
less interpretable than statistical models. Bridging the gap between the statistical and machine learning fields, we propose a novel Interpretable Neural Network Regression (INNER), which combines the strengths of statistical and DNN models. We use the proposed INNER to conduct individualized risk assessment of preoperative opioid use. Intensive simulations and an analysis of 34,186 patients expecting surgery in the Analgesic Outcomes Study (AOS) show that the proposed INNER not only can accurately predict the preoperative opioid use using preoperative characteristics as DNN, but also can estimate the patient-specific odds of opioid use without pain and the odds ratio of opioid use for a  unit increase in  the reported overall body pain, leading to more straightforward interpretations of the tendency to use opioids than DNN. Our results identify the patient characteristics that are strongly associated with opioid use and is largely consistent with the previous findings,  providing evidence that INNER is a useful tool for  individualized risk assessment of preoperative opioid use.
\end{abstract}


\section{Introduction}
\label{sec:intro}

The drastic increase in the use of opioids has led to an epidemic in the U.S., with more than 46,000 estimated overdose deaths  in 2018 [\cite{brown2021number}]. As an effort to combat this crisis, researchers have begun to study preoperative opioid use because it is a major factor associated with opioid  misuse~[\cite{saha2016nonmedical}], 
 higher postoperative  opioid demand~[\cite{armaghani2014preoperative,schoenfeld2018sustained}], worse postoperative outcomes~[\cite{lee2014preoperative,smith2017impact,morris2016preoperative,jain2018chronic}], and increased postoperative healthcare utilization and expenditures~[\cite{cron2017preoperative,waljee2017effect,jain2018chronic}]. Understanding preoperative opioid use among patients expecting surgical services can help surgeons establish effective pain management for  patients~[\cite{hilliard2018prevalence}],
including postoperative opioid management ~[\cite{roeckel2016opioid}].

What has often been overlooked is that a sizeable portion of patients
consumed opioids preoperatively even with no reported
pains ~[\cite{roeckel2016opioid}], which might hint at possible opioid misuse. 
As part of the Analgesic Outcomes Study (AOS) ~[\cite{brummett2013survey}],  a large observational cohort study investigating  associations between preoperative pain and opioid use,
individualized risk of preoperative opioid use   is assessed to identify  patients
who tend to use preoperative opioids even when there is little pain as well as those  who tend to take preoperative opoids even when the pain increases only slightly ~[\cite{brummett2015characteristics}]. With  opioid use (yes or no) as the outcome and  pain level as the covariate, a logistic regression model with
an intercept and a slope that depend on patients' other characteristics may help delineate the subgroups of patients who are at high risks of opioids misuse; see model (\ref{eq:INNER}). However, because of the curse of dimensionality, traditional nonparametric methods of fitting varying coefficient 
logistic models may not fare well~[\cite{park2015varying,hastie1993varying,cai2000efficient}], even when the number of patient characteristics is only moderately large.


On the other hand, deep neural network (DNN), 
 a  machine learning algorithm  inspired by the structure of  brains, has achieved much success in nonparametric approximation  with high dimensional predictors~[\cite{bauer2019deep,schmidt2020nonparametric}]. 
 It has found applications in  computational phenotyping~[\cite{che2015deep,lipton2015learning}], medical imaging analysis~[\cite{kleesiek2016deep,tran2016fully}] and predictive modeling~[\cite{choi2016doctor}], among many others. It is challenging to explain the decision rules of DNN with the input variables, due to the black-box nature;  directly applying DNN to the aforementioned AOS data cannot pinpoint the subgroup of patients who may be at high risks of opioid misuse. 

For example, \cite{dong2019machine} use the Gini importance index from the random forest model to rank features, while \cite{lo2019evaluation} use the 
boosting decision tree. However, neither of these methods can give the direction of the association between features and opioid dependence.
On the other hand, \cite{che2017deep} use a weight matrix of each layer in the DNN model to generate ``importance scores'' to detect important features. The scores not only rank different features in terms of opioid overdose prediction but also inform the direction of the association. However, the method may not directly decipher the relationship between opioid use and pain, or, in particular, identify subpopulations who are likely to be sensitive to pain or 
be opioid dependent even without reported pains.

Bridging the gap between the statistical and machine learning fields, we propose an interpretable neural network regression (INNER) that combines the strengths of logistic regression and DNN models.  We propose a logistic regression model with individualized coefficients, wherein  the regression coefficients are functions of individual characteristics. We utilize DNN to estimate these individualized coefficients and construct two metrics, Baseline Opioid Tendency (BOT) and Pain-induced Opioid Tendency (POT), which are useful for the individualized assessment of opioid use for each patient. In particular,   BOT refers to the odds of using preoperative opioids when the patient does not report pain and POT is the odds ratio of  using preoperative opioids for a unit increase in the reported overall body pain. These two metrics can be used to identify subgroups of patients, whose characteristics are associated with preoperative opioid use: patients with high POT are more likely  to get preoperative opioids when pain increases, and patients with high BOT have a high risk of preoperative opioid use even with no reported pain.
To demonstrate the utility of our proposal, we conduct simulations and apply the INNER model to analyze the AOS study.
Our analysis identifies patient characteristics that are associated with  opioid tendency, as quantified by BOT and POT,  and is largely consistent with the literature,  evidencing the usefulness of INNER  for  individualized risk assessment of preoperative opioid use.

\section{Review of a Deep Neural Network}
A DNN  has multiple layers with neurons being the basic processing units~[\cite{lecun2015deep}]. For example,  in the commonly used feedforward neural network~[\cite{shrestha2019review}], starting from the first layer (input layer), neurons in one layer are connected to and may ``activate'' those in the adjacent and higher layers. Specifically, the inputs of each neuron are multiplied by some weights,  added with respective bias terms and  summed up~[\cite{shrestha2019review,fan2021selective}]. The sums are  passed onto some transformation functions, called  ``activation'' functions, such as linear, Sigmoid, hyperbolic tangent or rectified linear unit (ReLU) activation functions~[\cite{karlik2011performance,li2017convergence}]. The outputs returned by these activation functions are fed to neurons in the next layer as inputs. Passing all of the layers, the outputs of the final layer (output layer) will be used for prediction.

\section{Interpretable Neural Network Regression}
Our proposed INNER model  is a logistic regression model with covariate-dependent coefficient functions constructed by DNN.  The general formulation is similar to that of a DNN. Specifically, let $\mathbb{R}^d$ be a $d$-dimensional Euclidean vector space. To construct a prediction  based on input  $\mathbf{x}\in \mathbb{R}^{k_{l}}$ 
via a neural network with $L$ layers, where the $l$th ($l = 1,\ldots, L$) layer consists of $k_l$ neurons, we adopt an $L$-fold composite function $F_L: \mathbb{R}^{k_1} \to \mathbb{R}^{k_{L+1}}$ with the parameter $\boldsymbol{\theta}$, i.e., 
$$F_L(\cdot;\boldsymbol{\theta}) = f_L \circ f_{L-1} \circ \cdots \circ f_{1}(\cdot),$$  where $f_l(\mathbf{x}) = \sigma_l(\mathbf{W}_l\mathbf{x} +\mathbf{b}_l) \in \mathbb{R}^{k_{l+1}}$ and ``$\circ$'' indicates the composition of two functions. The function $\sigma_l: \mathbb{R}^{k_{l+1}}\to \mathbb{R}^{k_{l+1}}$ is a (non)linear activation function for the $l$th layer. The parameter $\boldsymbol{\theta} = \{\mathbf{W}_l,\mathbf{b}_l\}_{l=1}^L$, where $\mathbf{W}_l$ is the weight matrix of dimension $k_{l+1}\times k_{l}$ and $\mathbf{b}_l \in \mathbb{R}^{k_{l+1}}$ is the bias vector. Typical choices of $\sigma_l(\mathbf{x})$ include a linear function of $\mathbf{x}$, a ReLU function, i.e., $\mathrm{max}(0,\mathbf{x})$,  and a softmax function, i.e., $\exp(\mathbf{x})/\|\exp(\mathbf{x})\|_1$, where  $\mbox{max}$ and $\exp$ operate componentwise. 

\begin{figure}[h!]
    \centering
    \includegraphics[scale=0.4]{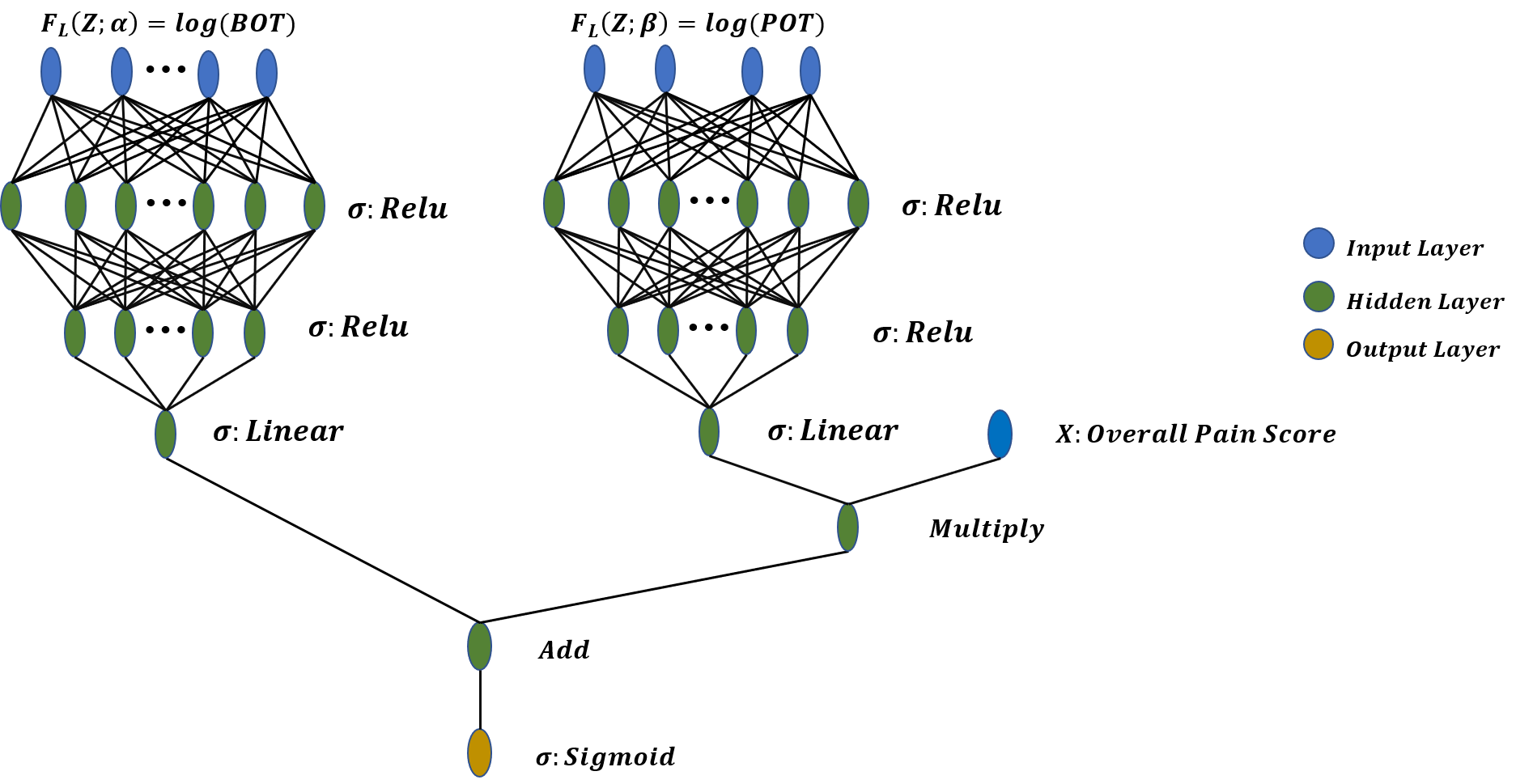}
    \caption{\textbf{Example of INNER.} \textbf{Input:} overall pain score (X) and other characteristics ($\mathbf{Z}$).  
    \textbf{Two neural networks for $F_L(\mathbf{Z};\boldsymbol{\alpha})$ and $F_L(\mathbf{Z};\boldsymbol{\beta})$:}  the same network architecture with different parameters, $\boldsymbol{\alpha}$ and $\boldsymbol{\beta}$;  three hidden layers in each  network, with the first layer having 250 neurons with a ReLu activation function, the second layer having 125 neurons with a ReLu activation function, and the last layer having one neuron with a linear activation function. \textbf{Ouput:} estimated probability of preoperative opioid use.  }
    \label{Fig3}
\end{figure}

Let
$
    \mathcal{D} = \{ ( X_i,\mathbf{Z}_i,Y_i),
    i=1, \ldots, N \}
$
be a dataset consisting of $N$ independent patients. For patient $i  \in \{1,\ldots, N\}$, let $Y_i \in\{0,1\}$ be a binary  variable indicating whether the patient uses  opioids preoperatively. Let $X_i\in [0,10]$ represent the overall body pain score and $\mathbf{Z}_i\in \mathbb{R}^p$ represent a vector of $p$ preoperative characteristics. We model the conditional probability of preoperative opioid use given the preoperative characteristics and the overall body pain score via
\begin{align}
    \mathrm{logit}\{\mathrm{P}(Y_i=1 \mid X_i,\mathbf{Z}_i)\}  = F_{L}(\mathbf{Z}_i;\boldsymbol{\alpha}) + F_{L}(\mathbf{Z}_i;\boldsymbol{\beta})\cdot X_i,\label{eq:INNER}
\end{align}
where $\mathrm{logit}(p) = \log\{p/(1-p)\}$, with  $p\in (0,1)$, is the logit link function. The two covariate-dependent coefficient functions  are constructed by two neural networks with the same network architecture but different parameters: $\boldsymbol{\alpha}$ and $\boldsymbol{\beta}$. Model (\ref{eq:INNER})
is termed an INNER model, wherein  the number of neurons in the input layer is $k_1 = p$ and the output layer  has only one neuron, i.e., $K_{L+1} = 1$. Figure \ref{Fig3} shows an example of three layers ($L = 3$), where the first two layers have 250 and 125 hidden neurons, respectively,  with a ReLU activation function,  and the third layer has one hidden neuron with a linear activation function. During training, we may randomly select a certain number of neurons in a layer and  ignore them in order to overcome overfitting [\cite{srivastava2014dropout}]. The proportion of
such ignored neurons in a layer is called the dropout rate with that layer. During testing, dropout is set to be inactive with  no neurons ignored.

We use the Sigmoid activation function, i.e., $\mathrm{Sigmoid}(x) = \{1 + \exp(-x)\}^{-1},$ for the  output layer of the INNER model.   Here,  $x$ comes from the affine combination of the two sub-networks whose final layers have a linear activation function, and the Sigmoid  function returns a value  between 0 and 1, ensuring numerical stability. The number of hidden layers, along with the number of hidden neurons and the dropout rate in each layer, are hyperparameters to be selected based on the prediction performance.

Our proposed INNER is interpretable within the traditional logistic regression framework, and can assess the individualized risk of preoperative opioid use
via two derived metrics: Baseline Opioid Tendency (BOT), the odds of taking opioid with no reported pain, and Pain-induced Opioid Tendency (POT), the odds ratio of taking opioid for a unit increase in overall body pain. In particular, BOT and POT can be represented by the output of the two neural networks with the input $\boldsymbol{Z}$ respectively:
\begin{align*}
    &\mbox{Baseline Opioid Tendency (BOT)}\coloneqq \mbox{exp}\{F_{L}(\mathbf{Z};\boldsymbol{\alpha})\},\\
    &\mbox{Pain-induced Opioid Tendency (POT)}\coloneqq \exp\{F_{L}(\mathbf{Z};\boldsymbol{\beta})\}.
\end{align*}
Therefore, a high Baseline Opioid Tendency (BOT) or a high Pain-induced Opioid Tendency (POT) indicates a potential high risk of taking preoperative opioids.

The estimates of parameters, denoted by $\hat{\boldsymbol{\alpha}}$ and $\hat{\boldsymbol{\beta}}$, are obtained by minimizing the negative log likelihood or the cross entropy loss function
\begin{align}
    \mathcal{L}(\boldsymbol{\alpha},\boldsymbol{\beta};\mathcal{D}) 
    &=-\sum_{i=1}^N Y_i \log\{\mathrm{P}(Y_i = 1 \mid X_i,\mathbf{Z}_i) \} + (1 - Y_i)\log\{1- \mathrm{P}(Y_i = 1 \mid X_i,\mathbf{Z}_i)\},
\end{align}
where $\mathrm{P}(Y_i = 1 \mid X_i,\mathbf{Z}_i)$ is as defined in (\ref{eq:INNER}).
We use stochastic gradient descent (SGD)~[\cite{bottou2010large}] for optimization. In our later data analysis and 
out of a total of 34,186 patients, we randomly assign 23,931 (70\%) patients to be training samples~($\mathcal{T}$) and the rest 10,256 (30\%) patients to be validation samples~($\mathcal{V}$) when computing the training  and validation loss.\\
\begin{algorithm}[h!]
\SetAlgoLined
\KwInput{learning rate ($\eta$), maximum difference ($\Delta$), batch size ($M$)}
\KwOutput{$\hat{\boldsymbol{\alpha}}$, $\hat{\boldsymbol\beta}$}
 \KwData{ Partition full data $\mathcal{D}$ to training  ($\mathcal{T}$) and validation ($\mathcal{V}$) samples}
 Initialization $\boldsymbol{\alpha}^0$, $\boldsymbol{\beta}^0$, training loss = validation loss\\
 \While{\upshape validation loss - training loss $\leq$ $\Delta$} {
        \For{\upshape mini-batch $m\gets1$ \KwTo $M$}{
        Draw random samples without replacement
        $(X_i,\mathbf{Z}_i,Y_i) \in \mathcal{T}$
    }
    Compute gradients $\nabla_{\boldsymbol{\alpha}}\mathcal{L}$ and $\nabla_{\boldsymbol{\beta}}\mathcal{L}$ of mini-batch\\
    Update parameters $\boldsymbol{\alpha} = \boldsymbol{\alpha}- \eta \nabla_{\boldsymbol{\alpha}}\mathcal{L}$ and $\boldsymbol{\beta} = \boldsymbol{\beta}- \eta \nabla_{\boldsymbol{\beta}}\mathcal{L}$\\
     Compute training loss: $\mathcal{L}(\boldsymbol{\alpha},\boldsymbol{\beta};\mathcal{T}) $\\
     Compute validation loss: $\mathcal{L}(\boldsymbol{\alpha},\boldsymbol{\beta};\mathcal{V}) $\\
 }
 \caption{Stochastic Gradient Descent}
\end{algorithm}

In general, classical stochastic gradient descent is sensitive to the choice of learning rates; a large learning rate gives fast convergence but  may induce numerical instability~[\cite{Liu2020On,darken1992learning}], while a small learning rate may ensure stability, though at the price of more iterative steps. In our implementation, we use grid search to tune the learning rates. For the real data analysis, we tune the learning rate over the range between 0.005 to 0.1 with 20 equally spaced grid points, and set  the batch size  to be 64  and the maximum difference between the training and validation loss to be $10^{-2}$. We obtain the estimates, $\hat{\boldsymbol{\alpha}}$ and $\hat{\boldsymbol{\beta}}$, after 200 iterations. We also conduct sensitivity analysis to assess the robustness of SGD towards the choices of these hyperparameters, and find the model's predictiveness performance is fairly robust to them; see Appendix B. 

With $\hat{\boldsymbol{\alpha}}$ and $\hat{\boldsymbol{\beta}}$, BOT and POT can be estimated by plugging in these estimates:  for a patient with $\mathbf{Z}_i$, the estimated BOT and POT are $\mbox{exp}\{F_{L}(\mathbf{Z}_i;\hat{\boldsymbol{\alpha}})\}$ and  $\mbox{exp}\{F_{L}(\mathbf{Z}_i;\hat{\boldsymbol{\beta}})\}$, respectively.

\section{Simulation Study}
We compare the prediction power and robustness of the proposed INNER with the existing methods, including decision trees, random forests, Bayesian additive regression trees regression (BART), support vector machine (SVM), logistic regression and DNN. Under various scenarios examined, we find that INNER outperform these competing methods. The prediction power of INNER is similar to or even better than DNN when the model assumptions of INNER hold, whereas  INNER  achieves a performance comparable to DNN even when the model assumptions are violated. Codes for the simulation study are provided in the Supplementary Material.

\subsection{Prediction Power}
We simulate data from a logistic regression model with non-linear varying-coefficient functions: 
\begin{align}
    &\mathrm{logit}\{\mathrm{P}(Y=1\mid X,  \mathbf{Z})\}
        = \sin(\mathbf{Z}^{\top}\boldsymbol{\alpha}) + \cos(\mathbf{Z}^\top\boldsymbol{\beta})\cdot X. \nonumber
\end{align}
The simulation study is designed with varying signal strengths, noise variances, number of covariates and sample sizes. The signal strength is measured by a signal-to-noise ratio, i.e.,
\begin{align*}
\frac{\mathrm{Var}\{\mathrm{P}(Y = 1\mid X,\mathbf{Z})\}}{\mathrm{Var}(Y) - \mathrm{Var}\{\mathrm{P}(Y = 1\mid X,\mathbf{Z})\}}.
\end{align*}
We assess the prediction power of INNER by varying the signal-to-noise ratio to be 0.2, 0.8 or 3.2, and setting the sample size and the number of signal covariates  to be 40,000 and 16, respectively. 

We next increase the noise variance by adding various numbers of noise covariates (8, 12 and 16)  into the data,  while fixing the signal-to-noise ratio, the number of samples and the number of covariates at 3.2, 40,000 and 16 respectively.  We finally consider several combinations of the numbers of covariates (9, 16, 32) and  samples (5,000; 10,000; 20,000), with a signal-to-noise ratio of 3.2 and in the absence of noise covariates. 

For each simulation configuration, we conduct a total of 500 experiments. In each experiment, we randomly allocate 80\% of the samples to the training data and the rest to the testing data, and compare seven methods: the INNER model [equation~\eqref{eq:INNER}], DNN, decision trees, random forests, BART and SVM models with combined $\mathbf{Z}$ and $X$ as the input, and the logistic regression model with a two-way interaction between $\mathbf{Z}$ and $X$. For the logistic regression, we present it as a special case of INNER with only one layer and a linear activation function, that is,

\begin{align}
    \mathrm{logit}\{\mathrm{P}(Y=1\mid X, \mathbf{Z})\} = \mathbf{Z}^\top\mathbf{W}_{\alpha} +b_{\alpha} + (\mathbf{Z}^\top\mathbf{W}_{\beta} +b_{\beta})\cdot X, \label{eq:logit_model}
\end{align}
where $\mathbf{W}_{\alpha}$ and $\mathbf{W}_{\beta}$ are the weight parameters, and $b_{\alpha}$ and $b_{\beta}$ are the bias terms.  

For INNER, the number of hidden layers in the neural network for $F_L(\mathbf{Z}_i;\boldsymbol{\alpha})$ is set to be 3. The first two layers have 200 and 10 hidden neurons with a dropout rate of 0.5 and 0.3, respectively.
These two layers are equipped with a ReLu activation function.  The final layer has only one neuron with a linear activation function. The neural network for $F_L(\mathbf{Z}_i;\boldsymbol{\beta})$ is similar and has 3  layers, each with 100, 90 and 1 hidden neurons  but
with no dropouts. The learning rate for both networks is 0.0014. 
For DNN, we use a network architecture with 4  layers: the first 3  layers have  160, 120 and 160 hidden neurons, respectively, and ReLu activation functions; the first and the third  layers are  with a dropout rate of 0.1 and 0.3, respectively; the last  layer has one hidden neuron and a Sigmoid activation function. 
The loss function and the optimizer are the same as in the INNER model, but with a learning rate of 0.0007.
These network hyperparameters are chosen to yield good prediction performances under the specified simulation configurations.

For decision trees, the maximum depth of a tree is 10 and the minimum number of samples required to split an internal node is 2. The minimum number of samples required to be at a leaf node for the decision is 4. For random forests, the number of features considered for the best split is the square root of the number of features, the minimum number of samples required to be at a leaf node is 2, the minimum number of samples required to split an internal node is 2 and the number of trees in a forest is 1,000. For SVM, we use a radial basis function kernel and set the regularization parameter to be 10. For BART, the number of trees to be grown in a sum-of-trees model is 80.

Summarizing the results of 500 simulations for each setting, Table \ref{CorrectModel}  shows that most of the models achieve better model performances as the signal-to-noise ratio increases.  For example, the C-statistics of decision trees and BART increase from 0.5 to more than 0.6 when the signal-to-noise  ratio increases from 0.2 to 3.2, while the C-statistic increases to more than 0.8 for random forests and SVM. The performance of INNER and DNN is comparable across different signal strengths and is better than that of the other models. The C-statistics of DNN and INNER are 0.96 when the signal-to-noise ratio is 3.2. 

Moreover, the performances of all the models deteriorate  with more noise covariates added (Table \ref{CorrectModel}). For instance, the C-statistic of BART decreases to around 0.6 with 16  added noise covariates, while the C-statistics for random forests and SVM, though slightly better, decrease to around 0.7 when we add 16 noise covariates. In contrast, the performances of INNER and DNN are consistently better than those of the other models. Moreover,  INNER  slightly outperforms DNN with noise covariates added;  with 16 added noise covariates, INNER achieves a C-statistic of 0.95, slightly better than 0.93 achieved by DNN.

With various combinations of the number of covariates and sample size, the performance of each model improves when we use more samples to train the model or decrease the number of covariates (Table \ref{CorrectModel}). DNN, INNER, random forests, BART and SVM achieve a C-statistic of more than 0.9 when the number of covariates is 8. Moreover, the C-statistics of DNN and INNER are 0.97 with 20,000 samples. However, when the number of covariates is 18, only random forests, SVM, DNN and INNER can achieve a  C-statistic of more than 0.6 with 20,000 samples. Also, INNER performs much better than DNN with smaller sample sizes and larger numbers of covariates. The C-statistic of INNER is 0.92, larger than 0.84 for DNN when the number of covariates is 18 and the  sample size is 20,000.
\begin{table}[ht]
\centering
\addtolength{\leftskip} {-2.5cm}
\addtolength{\rightskip}{-2.5cm}
\begin{threeparttable}
\caption{\label{CorrectModel}
Average (SE)  C-statistics for different methods under the correctly specified model}
\begin{tabular}{cccccccc}
\hline
 \backslashbox{\textbf{Specifications}}{\textbf{Model}}& \thead{Decision\\ Trees} & \thead{Random \\ Forests} & \thead{BART} & \thead{SVM} & \thead{Logistic \\ Regression} & \thead{DNN} & \thead{INNER} \\ 
 \hline
\textbf{Signal-to-noise Ratio$^a$} &  &  & & & & & \\ 
0.2 & 0.51 (0.0003) & 0.52 (0.0003) & 0.51 (0.0003) & 0.61 (0.0002) & 0.50 (0.0003) & 0.62 (0.0020) & 0.64 (0.0003)\\
0.8 & 0.62 (0.0003) & 0.73 (0.0002) & 0.61 (0.0004) & 0.75 (0.0002) &  0.50 (0.0009) & 0.84 (0.0002) &  0.85 (0.0002)\\
3.2 & 0.65 (0.0003) & 0.81 (0.0002) & 0.69 (0.0003) & 0.85 (0.0002) & 0.50 (0.0002) & 0.96 (0.0013) & 0.96 (0.0002)\\
\hline
\textbf{Noise Covariates$^b$} &  &  & & & & & \\ 
8 & 0.64 (0.0003) & 0.76 (0.0003) & 0.68 (0.0003) & 0.73 (0.0002) & 0.50 (0.0002) & 0.93 (0.0040) & 0.96 (0.0003)\\
12 & 0.64 (0.0003) & 0.74 (0.0003) & 0.67 (0.0003) & 0.73 (0.0002) & 0.50 (0.0002) & 0.93 (0.0036) & 0.96 (0.0003)\\
16 & 0.64 (0.0003) & 0.73 (0.0003) & 0.66 (0.0003) & 0.70 (0.0002) & 0.50 (0.0002) & 0.93 (0.0032) & 0.95 (0.0019)\\
\hline
\textbf{Number of Covariates$^c$} &  &  & & & & & \\
\textbf{8} &&&&&&&\\
Number of Samples &  &  & & & & & \\
5,000& 0.60 (0.0008)& 0.93 (0.0003) & 0.90 (0.0004) & 0.91 (0.0003) & 0.67 (0.0054) & 0.96 (0.0002) & 0.96 (0.0003)\\
10,000& 0.62 (0.0006)& 0.94 (0.0002) & 0.92 (0.0003) & 0.93 (0.0002) & 0.67 (0.0052) & 0.97 (0.0002) & 0.97 (0.0002)\\
20,000& 0.63 (0.0004) & 0.95 (0.0001) & 0.94 (0.0002) & 0.95 (0.0001) & 0.68 (0.0052) & 0.97 (0.0001) & 0.97 (0.0001)\\
\textbf{16}&&&&&&&\\
Number of Samples &  &  & & & & & \\
5,000& 0.60 (0.0008) & 0.68 (0.0008) & 0.59 (0.0007) & 0.74 (0.0006) & 0.50 (0.0005) & 0.89 (0.0005) & 0.88 (0.0010)\\
10,000& 0.62 (0.0006) & 0.72 (0.0005) & 0.62 (0.0005) & 0.78 (0.0004) &  0.50 (0.0004) & 0.92 (0.0004) & 0.93 (0.0005)\\
20,000 & 0.63 (0.0004) & 0.77 (0.0004) & 0.66 (0.0004) & 0.82 (0.0003) & 0.50 (0.0003) & 0.95 (0.0009) & 0.96 (0.0003)\\
\textbf{18}&&&&&&&\\
Number of Samples &  &  & & & & & \\
5,000 & 0.60 (0.0008) & 0.51 (0.0007) & 0.50 (0.0007) & 0.60 (0.0007) & 0.50 (0.0005) & 0.63 (0.0036) & 0.70 (0.0048)\\
10,000 & 0.62 (0.0006) & 0.54 (0.0008) & 0.51 (0.0006) & 0.62 (0.0004) & 0.50 (0.0003) & 0.78 (0.0046) & 0.86 (0.0036)\\
20,000 & 0.63 (0.0004) &  0.60 (0.0008)  & 0.51 (0.0006) & 0.66 (0.0003) & 0.50 (0.0002) & 0.84 (0.0057) & 0.92 (0.0049)\\
\hline
\end{tabular}
\begin{tablenotes}
\item[a.]the numbers of samples and covariates are fixed at 40,000 and 16, with varying signal-to-noise ratios and  no noise covariates
\item[b.]the signal-to-noise ratio, the numbers of samples and  covariates are fixed at 3.2, 40,000 and 16, with varying numbers of noise covariates
\item[c.]the signal-to-noise ratio is fixed at 3.2, with varying  numbers of covariates  and  samples and no noise variables
\end{tablenotes}
\end{threeparttable}
\end{table}

\subsection{Robustness}
We assess the robustness of INNER  when the INNER model
(\ref{eq:INNER}) deviates from the true data-generating  model, which is
\begin{equation}
\mathrm{logit}\{\mathrm{P}(Y = 1\mid X, \mathbf{Z})\} = -X \cdot \sin(\mathbf{Z}^{\top}\boldsymbol{\alpha}) +\sqrt{ |\cos(\mathbf{Z}^\top\boldsymbol{\beta})\cdot X|}. \nonumber
\end{equation}
The model structures of DNN and INNER used here differ from those in the prediction power study. DNN has four  layers: the first two layers each have 100 hidden neurons with a ReLu activation function; the third  layer has 160 neurons with a ReLu activation function and a dropout rate of 0.3; the last layer has one neuron with a Sigmoid function and a learning rate of 0.00046. For INNER, there are 3 hidden layers in the  neural networks of $F_L(\mathbf{Z}_i;\boldsymbol{\alpha})$ and $F_L(\mathbf{Z}_i;\boldsymbol{\beta})$. There are 200, 10 and 1  neurons in each layer of $F_L(\mathbf{Z}_i;\boldsymbol{\alpha})$, and 180, 90 and 1 neurons in each layer of $F_L(\mathbf{Z}_i;\boldsymbol{\beta})$.  The learning rate is set to be 0.004. Decision trees used here have a similar structure as those in the prediction power study, except that the minimum number of samples required to split an internal node is 10. For random forests, the maximum depth of a tree is 50, the minimum number of samples required to split an internal node is 2 and the number of trees in a forest is 2,500. For SVM, we use a radial basis function kernel and set the kernel coefficient to be 0.1. For BART, the number of trees to be grown in a sum-of-trees model is 90.

Based on 500 simulations for each setting,  Table \ref{Robustness} reveals that, even under a misspecified model, INNER is able to achieve a performance as good as DNN and continues to outperform the other models. For example,  when the signal-to-noise ratio is 3.2, both DNN and INNER achieve a C-statistic of 0.97, while the  C-statistics of all the other models are less than 0.9. With 16 noise covariates added, DNN and INNER  still achieve a C-statistic of 0.96, but the C-statistics for the other models are less than 0.8. When we increase the number covariates and decrease the number of samples, the C-statistics of DNN and INNER are still comparable and are higher than those of  the other models.

\begin{table}[ht]
\centering
\addtolength{\leftskip} {-2.5cm}
\addtolength{\rightskip}{-2.5cm}
\begin{threeparttable}
\caption{\label{Robustness}
Average (SE)  C-statistics for different methods under the misspecified model}
\begin{tabular}{cccccccc}
\hline
 \backslashbox{\textbf{Specifications}}{\textbf{Model}}& \thead{Decision\\ Trees} & \thead{Random \\ Forests} & \thead{BART} & \thead{SVM} & \thead{Logistic \\ Regression} & \thead{DNN} & \thead{INNER} \\ 
 \hline
\textbf{Signal-to-noise Ratio$^a$} &  &  & & & & & \\ 
0.2 & 0.55 (0.0003) & 0.60 (0.0003) & 0.55 (0.0003) & 0.61 (0.0002) & 0.50 (0.0004) & 0.70 (0.0003) & 0.71 (0.0020) \\
0.8 & 0.58 (0.0004) & 0.69 (0.0003) & 0.57 (0.0004) & 0.76 (0.0002) & 0.51 (0.0007) & 0.86 (0.0002) &  0.87 (0.0003)\\
3.2 & 0.62 (0.0003) & 0.78 (0.0003) & 0.62 (0.0005) & 0.82 (0.0002) & 0.51 (0.0006) & 0.97 (0.0002) & 0.97 (0.0003)\\
\hline
\textbf{Noise Covariates$^b$} &  &  & & & & & \\ 
8 & 0.61 (0.0003) & 0.76 (0.0003) & 0.61 (0.0005) & 0.71 (0.0002) & 0.51 (0.0005) & 0.96 (0.0002) & 0.96 (0.0003)\\

12 & 0.61 (0.0003) & 0.74 (0.0003) & 0.60 (0.0005) & 0.71 (0.0002) & 0.51 (0.0005) & 0.96 (0.0002) & 0.96 (0.0003)\\

16 & 0.61 (0.0003) & 0.73 (0.0003) & 0.59 (0.0005) & 0.68 (0.0002) & 0.51 (0.0005) & 0.96 (0.0002) & 0.96 (0.0010)\\

\hline
\textbf{Number of Covariates$^c$} &  &  & & & & & \\
\textbf{8} &&&&&&&\\
Number of Samples &  &  & & & & & \\
5,000&  0.55 (0.0009)& 0.93 (0.0003) & 0.91 (0.0004) & 0.91 (0.0004) & 0.57 (0.0054) & 0.94 (0.0004) & 0.94 (0.0006)\\
10,000& 0.58 (0.0007) & 0.94 (0.0002) & 0.93 (0.0003) & 0.93 (0.0002) &  0.57 (0.0056) &  0.96 (0.0002)   & 0.96 (0.0004)\\
20,000& 0.61 (0.0005) & 0.95 (0.0001) & 0.94 (0.0002) & 0.94 (0.0001) & 0.57 (0.0062) & 0.97 (0.0002) & 0.97 (0.0003)\\
\textbf{16}&&&&&&&\\
Number of Samples &  &  & & & & & \\
5,000 & 0.55 (0.0009) & 0.61 (0.0007) &  0.53 (0.0008) & 0.69 (0.0007) & 0.51 (0.0008) & 0.90 (0.0005) & 0.92 (0.0007)\\
10,000 & 0.58 (0.0007) & 0.68 (0.0005) & 0.56 (0.0007) & 0.75 (0.0004) &  0.51 (0.0007) & 0.94 (0.0003) & 0.95 (0.0004)\\
20,000 & 0.61 (0.0005) & 0.74 (0.0003) & 0.59 (0.0005) & 0.79 (0.0003) & 0.51 (0.0006) & 0.96 (0.0002) & 0.96 (0.0004)\\
\textbf{18}&&&&&&&\\
Number of Samples &  &  & & & & & \\
5,000 & 0.55 (0.0009) & 0.54 (0.0007)& 0.54 (0.0007) & 0.65 (0.0006) & 0.51 (0.0007) & 0.86 (0.0021) & 0.90 (0.0012)\\
10,000 & 0.58 (0.0007) & 0.53 (0.0005) & 0.54 (0.0005) & 0.68 (0.0004) &  0.51 (0.0005) & 0.92 (0.0005) & 0.94 (0.0014)\\
20,000 & 0.61 (0.0005) & 0.53 (0.0004) & 0.54 (0.0003) & 0.73 (0.0003) &  0.51 (0.0004) & 0.95 (0.0002) &  0.95 (0.0013)\\
\hline
\end{tabular}
\begin{tablenotes}
\item[a.]the numbers of samples  and  covariates are fixed at 40,000 and 16, with varying signal-to-noise ratios and no noise covariates
\item[b.]the signal-to-noise ratio, the numbers of samples and covariates are fixed at 3.2, 40,000 and 16, with varying numbers of noise covariates
\item[c.]the signal-to-noise ratio is 3.2, with varying numbers of covariates and  samples and no noise variables
\end{tablenotes}
\end{threeparttable}
\end{table}

\section{Analgesic Outcomes Study}
We use the proposed INNER model to study the associations between patient characteristics and preoperative opioid use.
\subsection{Data Preparation and Descriptive Analysis}
The data are  collected from the Analgesic Outcomes Study, an observational cohort study of acute and chronic pain~[\cite{brummett2017new,brummett2013survey,janda2015fibromyalgia,brummett2015characteristics,janda2015fibromyalgia,goesling2016trends}], with patients recruited  from the preoperative assessment clinic before the surgery or in the preoperative waiting area on the surgery day during daytime hours (approximately 5:30 AM to 5 PM). Patients are excluded if they do not speak English, are unable to provide written informed consent, or are incarcerated. The institutional review board of the University of Michigan, Ann Arbor, approved this study, and all participants provided written informed consent.  A total of 34,186 patients have been recruited and included in this analysis, and 7,894 (23.09\%) of them are identified to have used opioids at least once. Preoperative opioid use is dichotomized and used as the response variable in this study. Preoperative characteristics are collected using self-report measures of pain, function and mood. A total of 6,819 (19.95\%) patients have  missing values of preoperative characteristics, and we impute the missing data with the mean (for continuous variables) or the mode (for categorical variables). 
 
Sixteen preoperative characteristics are used to predict preoperative opioid use. Pain severity is measured with the Brief Pain Inventory~[\cite{tan2004validation}], which assesses overall, average and worst body pain (11-point Likert-type scale, with higher scores indicating greater pain severity). 
Briefly, among all the patients in the analysis, 54.2\% of them are female,  most of them are white (89.06\%), the mean age is 53.2 with a standard deviation (SD) of 16.2, and  7,984 (23.09\%) of these patients have taken  opioids preoperatively (Table \ref{Table1}). 

 It appears that some preoperative characteristics are associated with preoperative opioid use. Patients with more severe overall body pain (mean: 5.39, SD: 2.64) are more likely to use preoperative opioids. Smokers (4,341 [55.13\%]) are more likely to use preoperative opioids than non-smokers (10,142 [39.02\%]) (P $<$ 0.0001). Patients with illicit drug use history (614 [7.80\%]) and no alcohol consumption (4,677 [59.42\%]) are at higher risks of  preoperative opioid use (P $<$ 0.0001). Patients with anxiety (3,324 [47.98\%]), depression (2,409 [34.79\%]) or less satisfied with life (mean: 6.02, SD: 2.63) tend to use preoperative opioids (P $<$ 0.0001 for all). Patients who have poor physical conditions, e.g., those with American Society of Anaesthesiologists (ASA)  score of 3 or 4 (3,755 [47.57\%]) or high  Fibromyalgia Survey Score (mean: 8.34, SD: 5.25), are more likely to use preoperative opioids (P $<$ 0.001 for all). Preoperative opioid use is also associated with high BMI (mean: 30.74, SD: 7.74), sleep apnea (2,250 [29.15\%]), race (Asian: 39[0.49\%]) and  surgical type (P $<$ 0.0001 for all).
 
 \begin{table}[hbpt]
 \centering
\begin{threeparttable}
\caption{\label{Table1} Comparisons of Baseline Characteristics$^{a,b,c}$ }
\begin{tabular}{lcccr}
\hline
& \thead{Overall \\(N=34,186)} & \thead{Opioid Use \\(N=7,894)}& \thead{No Opioid Use \\(N=26,292)} &\textbf{P Value}\\ 
  \hline
BMI & 29.90 (7.18) & 30.74 (7.74) & 29.64 (6.98) & $<$0.0001 \\ 
  Age & 53.19 (16.15) & 53.37 (14.97) & 53.14 (16.49) & 0.2441 \\ 
   Fibromyalgia Survey Score & 5.47 (4.63) & 8.34 (5.25) & 4.61 (4.05) & $<$0.0001 \\ 
  Satisfaction with Life & 7.03 (2.57) & 6.02 (2.63) & 7.33 (2.47) & $<$0.0001 \\ 
  Charlson Comorbidity Index & 1.68 (3.30) & 1.65 (3.32) & 1.69 (3.29) & 0.3037 \\ 
  Overall BPI Score&3.21 (2.86)& 5.39 (2.64) &2.55 (2.58) &$<$0.0001\\
  Gender &  &  &  & 0.2605 \\ 
    \qquad Female & 18,530 (54.20) & 4,323 (54.76) & 14,207 (54.04) &  \\ 
    \qquad Male & 15,656 (45.80) & 3,571 (45.24) & 12,085 (45.96) &  \\ 
  Race &  &  &  & $<$0.0001 \\ 
   \qquad White & 30,445 (89.06) & 6,979 (88.41) & 23,466 (89.25) &  \\ 
   \qquad African American & 1,780 (5.21) & 529 (6.70) & 1,251 (4.76) &  \\ 
   \qquad Asian & 467 (1.37) & 39 (0.49) & 428 (1.63) &  \\ 
  \qquad Other & 1,494 (4.37) & 347 (4.40) & 1,147 (4.36) &  \\ 
  Tobacco use &  &  &  & $<$0.0001 \\ 
    \qquad No & 19,384 (57.24) & 3,533 (44.87) & 15,851 (60.98) &  \\ 
    \qquad Yes & 14,483 (42.76) & 4,341 (55.13) & 10,142 (39.02) &  \\ 
  Alcohol consumption &  &  &  & $<$0.0001 \\ 
    \qquad No & 18,755 (55.39) & 4,677 (59.42) & 14,078 (54.17) &  \\ 
    \qquad Yes & 15,105 (44.61) & 3,194 (40.58) & 11,911 (45.83) &  \\ 
  Illicit drug use &  &  &  & $<$0.0001 \\ 
    \qquad No & 32,382 (95.61) & 7,260 (92.20) & 25,122 (96.65) &  \\ 
    \qquad Yes & 1,486 (4.39) & 614 (7.80) & 872 (3.35) &  \\ 
  Sleep apnea &  &  &  & $<$0.0001 \\ 
    \qquad No & 25,210 (75.96) & 5,468 (70.85) & 19,742 (77.51) &  \\ 
    \qquad Yes & 7,977 (24.04) & 2,250 (29.15) & 5,727 (22.49) &  \\ 
  Depression &  &  &  & $<$0.0001 \\ 
    \qquad No & 24,278 (80.40) & 4,515 (65.21) & 19,763 (84.91) &  \\ 
    \qquad Yes & 5,920 (19.60) & 2,409 (34.79) & 3,511 (15.09) &  \\ 
  Anxiety &  &  &  & $<$0.0001 \\ 
    \qquad No & 19,368 (64.15) & 3,604 (52.02) & 15,764 (67.77) &  \\ 
    \qquad Yes & 10,822 (35.85) & 3,324 (47.98) & 7,498 (32.23) &  \\ 
  ASA Score &  &  &  & $<$0.0001 \\ 
    \qquad 1-2 & 21,898 (64.06) & 4,139 (52.43) & 17,759 (67.55) &  \\ 
    \qquad 3-4 & 12,288 (35.94) & 3,755 (47.57) & 8,533 (32.45) &  \\ 
  Body Group &  &  &  & $<$0.0001 \\ 
    \qquad Head & 3,714 (11.06) & 745 (9.52) & 2,969 (11.53) &  \\ 
    \qquad Neck & 4,150 (12.36) & 806 (10.30) & 3,344 (12.99) &  \\ 
    \qquad Thorax & 2,167 (6.45) & 363 (4.64) & 1,804 (7.01) &  \\ 
    \qquad Intrathoracic & 15,53 (4.62) & 244 (3.12) & 1,309 (5.08) &  \\ 
    \qquad Shoulder/Axilla & 1854 (5.52) & 321 (4.10) & 1,533 (5.95) &  \\
    \qquad Upper Arm \& Elbow & 245 (0.73) & 88 (1.12) & 157 (0.61) &  \\
    \qquad Forearm, Wrist, Hand & 1359 (4.05) & 348 (4.45) & 1,011 (3.93) &  \\ 
    \qquad Upper Abdomen & 3,298 (9.82) & 765 (9.77) & 2,533 (9.84) &  \\ 
    \qquad Lower Abdomen & 4,963 (14.78) & 962 (12.29) & 4,001 (15.54) &  \\
    \qquad Spine/Spinal Cord & 1,472 (4.38) & 841 (10.74) & 631 (2.45) &  \\ \qquad Perineum & 3497 (10.41) & 728 (9.30) & 2,769 (10.75) &  \\ 
    \qquad Pelvis (Except Hip) & 125 (0.37) & 53 (0.68) & 72 (0.28) &  \\
    \qquad Upper Leg (Except Knee) & 1,582 (4.71) & 567 (7.24) & 1,015 (3.94) &  \\ 
    \qquad Knee/Popliteal & 1,933 (5.76) & 401 (5.12) & 1,532 (5.95) &  \\ 
    \qquad Lower Leg & 772 (2.30) & 309 (3.95) & 463 (1.80) &  \\ 
    \qquad Other & 896 (2.67) & 287 (3.67) & 609 (2.36) &  \\ 
    \hline
\end{tabular}
\begin{tablenotes}
\item[a.] mean (SD) for each continuous characteristic is reported 
\item[b.] frequency (percentage) for each categorical characteristic is reported
\item[c.]$\chi^2$ test or unpaired 2-tailed t test is used to assess the univariate differences between non-users and opioid users  as appropriate
\end{tablenotes}
\end{threeparttable}
\end{table}

\subsection{Prediction Performance Evaluation}
We compare the prediction performance of INNER with that of DNN and the logistic regression. We randomly split the data into the training  and testing parts. Data imputation is then performed for the training  and testing data separately. After training the INNER model using the training data, we test the prediction performance on the testing data. We conduct 100 independent trials by repeating the same procedure. We compare INNER with DNN and the logistic regression in accuracy, C-statistic, sensitivity, specificity and balance accuracy (the average of sensitivity and specificity). 

Since the outcome data are unbalanced, we further propose a balanced subsampling strategy to avoid overfitting. That is, we split each training dataset into the opioid user and non-user groups.  Among the non-user group, we randomly select the same number of patients as in the user group, append them to the user group and form a ``balanced'' dataset. We repeat the same procedure five times,  generating five datasets and training  five models on them.  We then apply these five models to the testing data and compute the probability of taking opioids by averaging the probabilities estimated by these models. We also use different thresholds to predict whether a patient takes opioids. For example, the threshold can be  50.00\% or 23.09\%, the prevalence of taking opioids in the original data; patients with estimated probabilities of taking opioids higher than the threshold are predicted as using opioids.

The architecture of INNER is the same as the example shown in Figure \ref{Fig3}.  There are multiple hidden layers in each of the neural network, $F_L(\mathbf{Z}_i;\boldsymbol{\alpha})$ and $F_L(\mathbf{Z}_i;\boldsymbol{\beta})$. The last hidden layer has a linear activation function while the other layers  have a ReLu activation function. We tune the number of layers and the number of hidden neurons in each layer based on the metrics we mentioned above. The best architecture has three hidden layers, each with 250, 125 and 1 hidden neuron, respectively. 

When tuning the architecture of INNER, we vary the number of hidden layers and the number of neurons in each layer for $F_L(\mathbf{Z}_i;\boldsymbol{\alpha})$ and $F_L(\mathbf{Z}_i;\boldsymbol{\beta})$ and compare the different architectures based on accuracy, C-statistic, sensitivity, specificity and balance accuracy. We vary the number of hidden layers from 2 to 5, and the number of neurons in the first layer to be 125, 250 or 500. We set the number of neurons in the last layer to be 1. Each of the rest hidden layers has half of the previous layer's neurons. In Section B of the Supplementary Material, we show the performance of the best architecture and two other more complicated architectures.

For DNN, we define multiple inputs based on the nature of preoperative characteristics. We classify the characteristics into three categories, un-modifiable such as gender and race, modifiable such as BMI, alcohol and smoking status, and directly pain-related such as pain severity and  Fibromyalgia Survey Score.  As shown in Appendix A, each input is passed to the same structure of hidden layers by a ReLu activation function with different parameters, and is concatenated and passed to the output layer. We tune the number of hidden layers and the number of neurons in each layer before concatenation. The best architecture before concatenation has two layers and the number of hidden neurons in each of the corresponding layers is 500 and 250. After concatenation, there is one hidden layer with 15 neurons. The loss function and optimizer of DNN is the same as that of INNER.  In Section B of the Supplementary Material, we shows the performance of the best architecture and two other more complicated architectures.

The INNER model achieves similar prediction power as DNN and is better than the logistic regression. We assess the performance of the three models with the best architectures using different sampling strategies and different; see Section B of the Supplementary Material. All the three models achieve the best  balance accuracy with balance sampling and 0.5 as the threshold. Table \ref{Table2} report the best performance of three models. The INNER and DNN achieve  better performance in all four metrics (accuracy, sensitivity, specificity and balance accuracy) compared to the logistic regression. Moreover,  there is no significant difference between the performance of INNER~(accuracy: 0.72, SE: 0.0029; sensitivity: 0.69,  SE: 0.0052; specificity: 0.73, SE: 0.0052; and balance accuracy: 0.71, SE: 0.0008) and DNN (accuracy: 0.72, SE: 0.0017; sensitivity: 0.694, SE: 0.0043; specificity: 0.73, SE: 0.0034; and balance accuracy: 0.71, SE: 0.0007). 

\begin{table}[ht]
\centering
\begin{threeparttable}
\caption{\label{Table2}Comparisons of Model Goodness-of-fit with the AOS data$^{a,b}$}
\begin{tabular}{lccc}
\hline
 &\textbf{Deep Neural Network} &\textbf{Logistic Regression}&\textbf{INNER}\\ 
  \hline
C-statistic & 0.78 (0.0006) & 0.76 (0.0027) & 0.78 (0.0006)\\ 
Accuracy &  0.76 (0.0017) & 0.63 (0.0129) & 0.72 (0.0029) \\ 
Sensitivity &  0.69 (0.0043) & 0.67 (0.0261) & 0.69 (0.0052) \\ 
Specificity &0.73 (0.0034) & 0.62 (0.0238) & 0.73 (0.0052) \\ 
Balance Accuracy & 0.71 (0.0007) & 0.64 (0.0049) & 0.71 (0.0008) \\ 
\hline
\end{tabular}
\begin{tablenotes}
\item[a.] the results are obtained under the best architecture (for DNN and INNER) with the balanced subsampling strategy and a threshold of 0.5. For the performance of different sampling strategies, thresholds or structures, refer to Appendix B
\item[b.] based on 100 random splits.
 
\end{tablenotes}
\end{threeparttable}
\end{table}

\subsection{Subgroup Analysis}
We identify different subgroups based on the local false discovery rates~[\cite{efron2004large,efron2007size,efron2007correlation}] by looking at the distributions of the estimated  POT and BOT. We then perform descriptive analysis on each of the subgroups and report the means (standard deviations) for continuous preoperative characteristics, and the frequencies (percentages) for categorical preoperative characteristics. We also estimate the probability of taking preoperative opioids for each patient with different pain scores and plot the average probability of taking preoperative opioids for each subgroup.

\begin{figure}[ht]
    \centering
    \includegraphics[scale=0.4]{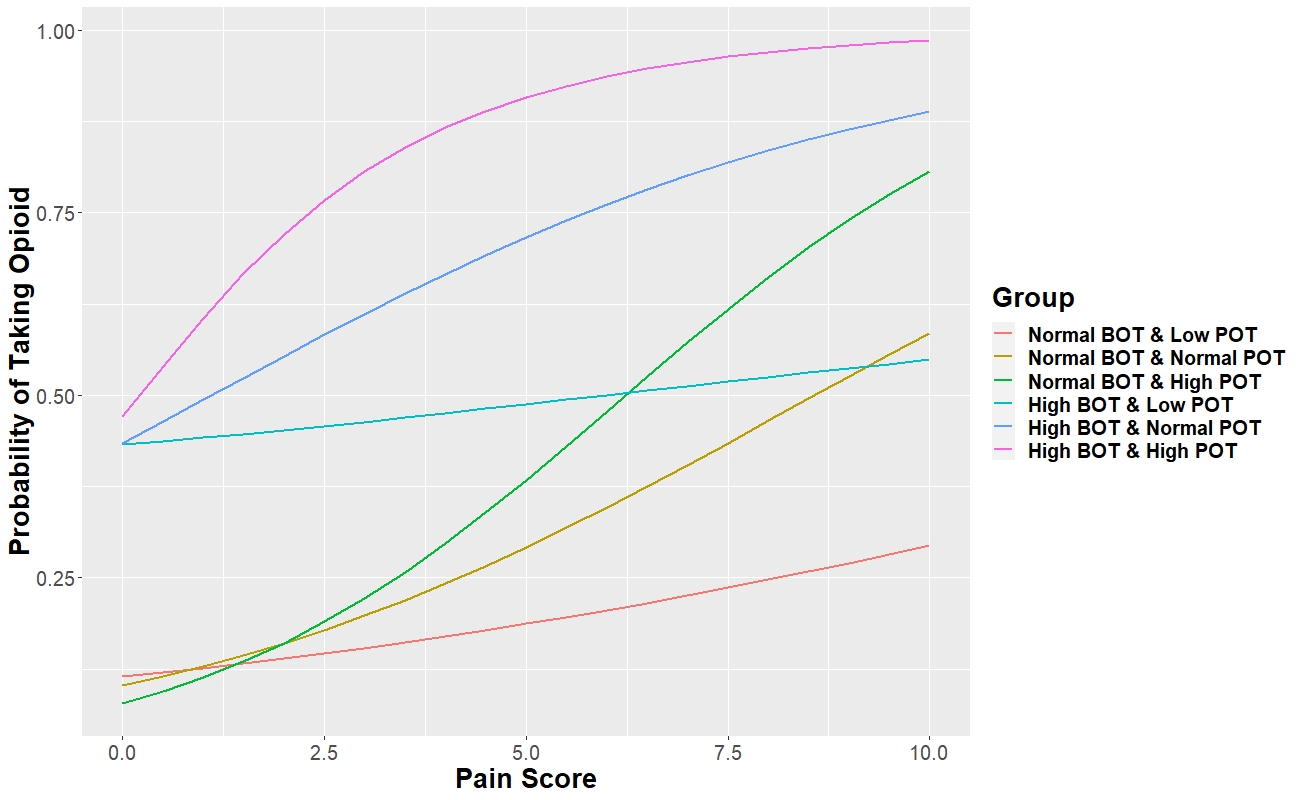}
    \caption{\textbf{Estimated Probability of Taking Preoperative Opioids Against Pain Score Stratified by Risk Groups} 
    } 
    \label{Fig2}
\end{figure}

 Our results lead to  six subgroups (Fig \ref{Fig2}) by controlling the local false discovery rate at 0.2. These subgroups include normal BOT \& low POT (4,889 patients), normal BOT \& normal POT (25,581 patients), normal BOT \& high POT (3,579 patients), high BOT \& low POT (67 patients), high BOT \& normal POT (47 patients) and high BOT \& high POT (6 patients). We estimate the probability of taking preoperative opioids with different pain scores stratified by subgroups in Fig \ref{Fig2}.  For the high BOT \& high POT subgroup and the high BOT \& normal POT subgroup, the probability of taking opioids exceeds 0.5 when the pain score is relatively low (high BOT \& high POT: 0.2; high BOT \& normal POT: 1.1), indicating these two subgroups have a high risk of taking preoperative opioids. The probability of taking opioids only exceeds 0.5 at the pain score of 6.0 for the high BOT \& low POT subgroup and 6.3 for the normal BOT \& high POT subgroup, and these two subgroups are considered as a moderate risk group. Finally, the normal BOT \& normal POT subgroup has probability of taking opioids higher than 0.5 only when the pain score is larger than 8.6, and the probability for normal BOT \& low POT is lower than 0.5 even when the pain score is 10. Thus, the normal BOT \& normal POT subgroup and normal BOT \& low POT subgroup are considered as a low risk group.

Tables \ref{SubgroupAna1} and  \ref{SubgroupAna2} show the characteristics for each subgroup. Patients in the high risk group (high BOT \& normal and high BOT \& high POT) and the moderate risk group (normal BOT \& high POT and high BOT \& low POT) tend to be more obese, younger, and have higher  Fibromyalgia Survey Scores and higher Charlson Comorbidity Indices than those in the low risk group. African Americans constitute about 17\% of patients in the high risk group,  while there are only 5\% African Americans in the low risk group. Most of patients (100\% for high BOT \& normal POT group and 83\% for high BOT \& high POT group) in the high risk group have tobacco consumption.  Patients are more likely to have illicit drug use history and sleep apnea in the high and moderate risk groups than in the low risk group. A large portion of patients in the high risk group (high BOT \& normal POT: 97.83\%, high BOT \& high POT: 66.67\%) have an ASA score of 3 or above,  indicating a very poor overall physical condition.

We also perform an ANCOVA-type analysis to understand the  importance of each covariate's  contributions to the developed risk scores  (Table \ref{SubgroupAna1}, Table \ref{SubgroupAna2}). Specifically, we use the log-transformed POT and BOT  as response variables to fit  separate linear models  and calculate $R^2$ for each preoperative characteristic. Based on the $R^2$, Fibromyalgia Survey Score and ASA Score explain the most variations of BOT, while Fibromyalgia Survey Score, age and Charlson Comorbidity Index explain the most variations of POT.
\begin{landscape}
\begin{table}
\caption{\label{SubgroupAna1}Subgroup Analysis$^a$}
\begin{tabular}{lcccccccc}
\hline
& \thead{Normal BOT \\ Low POT \\(N=4,889)} & \thead{Normal BOT\\Normal POT \\(N=25,581)}& \thead{Normal BOT\\High POT\\(N=3,597)} &\thead{High BOT\\Low POT \\(N=67)}&\thead{High BOT\\Normal POT \\(N=46)}&\thead{High BOT\\High POT \\(N=6)} & \thead{$\mathbf{R^2}$\\(BOT,\%)} & \thead{$\mathbf{R^2}$\\(POT,\%)} \\ 
  \hline
POT & 1.13 (0.07) & 1.31 (0.05) & 1.54 (0.12) & 1.05 (0.13) & 1.27 (0.06) & 1.79 (0.49) \\ 
  BOT & 0.14 (0.10) & 0.12 (0.08) & 0.09 (0.07) & 0.77 (0.14) & 0.78 (0.13) & 0.94 (0.38) \\ 
  BMI & 29.63 (7.28) & 29.62 (6.41) & 32.18 (10.31) & 33.99 (9.27) & 28.95 (8.80) & 36.44 (23.02) & 0.91 & 1.29\\ 
  Age & 52.27 (19.17) & 55.02 (14.74) & 41.68 (16.48) & 47.58 (10.61) & 50.2 (12.97) & 28.5 (6.98) &1.86 & 2.87\\ 
   Fibromyalgia Survey Score & 1.16 (3.20) & 5.45 (4.62) & 5.38 (4.43) & 2.1 (7.47) & 4.2 (9.55) & 8.83 (13.72) & 9.47 & 8.22\\ 
  Satisfaction with Life & 9.27 (1.45) & 6.99 (2.58) & 7.29 (2.75) & 9.6 (1.59) & 8.7 (2.84) & 8.5 (2.51) &2.89 & 1.18\\ 
  Charlson Comorbidity Index & 1.13 (3.07) & 1.63 (3.07) & 2.77 (4.55) & 3 (4.86) & 2 (4.08) & 6.67 (5.39) &0.06 & 2.75\\ 
  Gender &  &  &  &  &  &  & 0.60 & 0.08\\ 
    \qquad Female & 2,557 (52.30) & 13,854 (54.16) & 2,056 (57.16) & 38 (56.72) & 21 (45.65) & 4 (66.67) \\ 
    \qquad  Male & 23,32 (47.70) & 11,727 (45.84) & 1,541 (42.84) & 29 (43.28) & 25 (54.35) & 2 (33.33) \\ 
  Race &  &  &  &  &  &  &0.22 & 0.03\\ 
   \qquad  White & 4,266 (87.26) & 22,915 (89.58) & 3,171 (88.16) & 52 (77.61) & 36 (78.26) & 5 (83.33) \\ 
   \qquad  African American & 288 (5.89) & 1,263 (4.94) & 208 (5.78) & 12 (17.91) & 8 (17.39) & 1 (16.67) \\ 
   \qquad  Asian & 95 (1.94) & 327 (1.28) & 44 (1.22) & 1 (1.49) & 0 (0.00) & 0 (0.00) \\ 
  \qquad  Other & 240 (4.91) & 1,076 (4.21) & 174 (4.84) & 2 (2.99) & 2 (4.35) & 0 (0.00) \\ 
  Tobacco use &  &  &  &  &  &  & 16.96 & 0.69\\ 
   \qquad   No & 3,217 (65.80) & 14,417 (56.36) & 2,064 (57.38) & 4 (5.97) & 0 (0.00) & 1 (16.67) \\ 
   \qquad   Yes & 1,672 (34.20) & 11,164 (43.64) & 1,533 (42.62) & 63 (94.03) & 46 (100) & 5 (83.33) \\ 
  Alcohol consumption &  &  &  &  &  &  & 1.36 & < 0.01\\ 
   \qquad   No & 2,807 (57.41) & 14,026 (54.83) & 2,150 (59.77) & 56 (83.58) & 38 (82.61) & 4 (66.67) \\ 
   \qquad   Yes & 2,082 (42.59) & 11,555 (45.17) & 1,447 (40.23) & 11 (16.42) & 8 (17.39) & 2 (33.33) \\ 
  Illicit drug use &  &  &  &  &  &  & 0.97 & < 0.01\\ 
    \qquad  No & 4,718 (96.50) & 24,506 (95.80) & 3,380 (93.97) & 60 (89.55) & 32 (69.57) & 4 (66.67) \\ 
    \qquad  Yes & 171 (3.50) & 1,075 (4.20) & 217 (6.03) & 7 (10.45) & 14 (30.43) & 2 (33.33) \\ 
  Sleep apnea &  &  &  &  &  &  & 1.48 & 0.02\\ 
    \qquad  No & 3,948 (80.75) & 19,352 (75.65) & 2,843 (79.04) & 36 (53.73) & 25 (54.35) & 5 (83.33) \\ 
   \qquad   Yes & 941 (19.25) & 6,229 (24.35) & 754 (20.96) & 31 (46.27) & 21 (45.65) & 1 (16.67) \\ 
    \hline
\end{tabular}
\begin{tablenotes}
\item[a.] mean (SD) for each continuous characteristic  and frequency (percentage) for each categorical characteristic are reported. 
\end{tablenotes}
\end{table}
\begin{table}
\caption{\label{SubgroupAna2}Subgroup Analysis$^a$ (Continued)}
\begin{tabular}{lcccccccc}
\hline
& \thead{Normal BOT \\ Low POT \\(N=4,889)} & \thead{Normal BOT\\Normal POT \\(N=25,581)}& \thead{Normal BOT\\High POT\\(N=3,597)} &\thead{High BOT\\Low POT \\(N=67)}&\thead{High BOT\\Normal POT \\(N=46)}&\thead{High BOT\\High POT \\(N=6)} & \thead{$\mathbf{R^2}$\\(BOT,\%)} & \thead{$\mathbf{R^2}$\\(POT,\%)} \\ 
\hline
Depression &  &  &  &  &  &  & 1.50 & 0.29\\ 
    \qquad  No & 4,692 (95.97) & 20,444 (79.92) & 3,022 (84.01) & 63 (94.03) & 40 (86.96) & 5 (83.33) \\ 
    \qquad  Yes & 197 (4.03) & 5,137 (20.08) & 575 (15.99) & 4 (5.97) & 6 (13.04) & 1 (16.67) \\ 
Anxiety &  &  &  &  &  &  & 0.79 & 0.20\\ 
    \qquad  No & 4,383 (89.65) & 16,567 (64.76) & 2,308 (64.16) & 62 (92.54) & 40 (86.96) & 4 (66.67) \\ 
   \qquad   Yes & 506 (10.35) & 9,014 (35.24) & 1,289 (35.84) & 5 (7.46) & 6 (13.04) & 2 (33.33) \\ 
  ASA score &  &  &  &  &  &  & 8.26 & < 0.01\\ 
    \qquad  0-2 & 3,389 (69.32) & 16,102 (62.95) & 2402 (66.78) & 2 (2.99) & 1 (2.17) & 2 (33.33) \\ 
    \qquad  3-4 & 1,500 (30.68) & 9,479 (37.05) & 1,195 (33.22) & 65 (97.01) & 45 (97.83) & 4 (66.67) \\ 
  Body area &  &  &  &  &  &  &2.05 & 2.29\\ 
      \qquad  Head & 835 (17.08) & 2,507 (9.80) & 361 (10.04) & 9 (13.43) & 1 (2.17) & 1 (16.67) \\ 
          \qquad  Neck & 480 (9.82) & 3,198 (12.5) & 460 (12.79) & 5 (7.46) & 6 (13.04) & 1 (16.67) \\ 
       \qquad  Thorax & 286 (5.85) & 1,660 (6.49) & 214 (5.95) & 4 (5.97) & 3 (6.52) & 0 (0.00) \\ 
      \qquad  Intrathoracic & 332 (6.79) & 1,095 (4.28) & 120 (3.34) & 5 (7.46) & 1 (2.17) & 0 (0.00) \\ 
         \qquad  Shoulder/Axilla & 279 (5.71) & 1,399 (5.47) & 170 (4.73) & 5 (7.46) & 1 (2.17) & 0 (0.00) \\ 
            \qquad  Upper Arm \& Elbow & 18 (0.37) & 189 (0.74) & 36 (1.00) & 0 (0.00) & 2 (4.35) & 0 (0.00) \\ 
    \qquad  Forearm, Wrist, Hand & 414 (8.47) & 848 (3.31) & 94 (2.61) & 3 (4.48) & 0 (0.00) & 0 (0.00) \\ 
    \qquad  Upper Abdomen & 325 (6.65) & 2,443 (9.55) & 506 (14.07) & 16 (23.88) & 8 (17.39) & 0 (0.00) \\ 
    \qquad  Lower Abdomen & 716 (14.65) & 4,176 (16.32) & 666 (18.52) & 6 (8.96) & 3 (6.52) & 2 (33.33) \\ 
       \qquad  Spine/Spinal Cord & 66 (1.35) & 1,296 (5.07) & 102 (2.84) & 4 (5.97) & 4 (8.70) & 0 (0.00) \\ 
      \qquad   Perineum & 416 (8.51) & 2,707 (10.58) & 359 (9.98) & 7 (10.45) & 8 (17.39) & 0 (0.00) \\ 
       \qquad  Pelvis (Except Hip) & 12 (0.25) & 92 (0.36) & 20 (0.56) & 0 (0.00) & 1 (2.17) & 0 (0.00) \\ 
    \qquad   Upper Leg (Except Knee) & 107 (2.19) & 1352 (5.29) & 120 (3.34) & 0 (0.00) & 2 (4.35) & 1 (16.67) \\ 
    \qquad  Knee/Popliteal & 424 (8.67) & 1,384 (5.41) & 124 (3.45) & 0 (0.00) & 1 (2.17) & 0 (0.00) \\ 
    \qquad  Lower Leg & 89 (1.82) & 540 (2.11) & 140 (3.89) & 1 (1.49) & 1 (2.17) & 1 (16.67) \\ 
  \qquad  Other & 90 (1.84) & 695 (2.72) & 105 (2.92) & 2 (2.99) & 4 (8.7) & 0 (0.00) \\ 
  \hline
\end{tabular}
\begin{tablenotes}
\item[a.] mean (SD) for each continuous characteristic  and frequency (percentage) for each categorical characteristic are reported. 
\end{tablenotes}
\end{table}
\end{landscape}

\subsection{Discussion}
The proposed INNER model achieves predictability comparable to DNN, but with  more interpretability. The model leads to  two metrics, BOT and POT, that may decipher the patterns of preoperative opioid use and explain the association between preoperative characteristics and preoperative opioid use. Patients with higher BMI and worse physical conditions (higher Charlson Comorbidity Indices, higher  Fibromyalgia Survey Scores and higher ASA Scores) are more likely to consume preoperative opioids, and  African American patients are more likely to be in the high and moderate risk  groups. Patients with illicit drug use history and tobacco consumption are more likely to take preoperative opioids, while patients with alcohol consumption are less likely to have preoperative opioids. Patients with sleep apnea  have a higher risk of taking preoperative opioids, as do  patients expecting upper abdomen surgery. Detailed discussions  of these subgroups  can be found in the Appendix C.

Our results are largely consistent with the literature, which shows, for example, that patients with worse physical conditions are  more likely to use preoperative opioids~[\cite{sullivan2005regular,westermann2018epidemiology,goesling2015symptoms,meredith2019preoperative}]. \cite{prentice2019preoperative} find that high BMI and Black race are preoperative risk factors for opioid use. Younger patients are reported to have a higher risk of preoperative opioid use controlling for sociodemographics and clinical variables~[\cite{sullivan2006association}]. Similarly, \cite{lo2019evaluation} find that age is the among the most important features for opioid overdose prediction. \cite{hah2015factors} report that patients with poor sleep quality are more likely to have preoperative opioid use. Tobacco use is reported to be a risk factor of preoperative opioid use by many studies~[\cite{westermann2018epidemiology,meredith2019preoperative}]. As for substance abuse, many studies find that subjects with drug use have higher risks of opioid use~[\cite{sullivan2006association,sullivan2005regular}]. Both \cite{dong2019machine} and \cite{che2017deep} find that substance abuse history is among the most important features for opioid dependence prediction. \cite{sullivan2006association} find that there is no significant association between problem alcohol use and opioid prescription (OR: 0.63; 95\% CI: 0.35-1.15). The direction of OR in their study is consistent with our results. \cite{sullivan2005regular} report a non-significant association between alcohol use and opioid prescription, though the direction is opposite from our study (OR: 1.32, P = 0.479). More studies are warranted to identify the association between alcohol consumption and preoperative opioid use.

Finally, our proposed model can be extended to accommodate generalized linear models (GLMs) as discussed in \cite{tran2020bayesian}. Specifically,  let $g(\cdot)$ be a link function to link the conditional mean $\mathbb{E}(y\mid x)=g^{-1}\{\eta(x)\}$ to covariates of interest (e.g., treatment or exposure), say, $x$, where $\eta(x) = \beta_0 + \boldsymbol{\beta}^\top x$. In order to model the nonlinear effects of additional features $z$ (e.g., demographics, biomarkers)  on $\eta$ and achieve model flexibility, we can extend deep neural network to model the  individualized intercepts and coefficients, namely,  $\beta_0(z)$ and $\boldsymbol{\beta}(z)$. As such, the predictor can be written as $\eta(x,z) = \beta_0(z) + \boldsymbol{\beta}(z)^\top x$, which is to be linked to the conditional mean $\mathbb{E}(y\mid x,z) $ via $ \mathbb{E}(y\mid x,z)= g^{-1}\{\eta(x,z)\} = g^{-1}\{\beta_0(z) + \boldsymbol{\beta}(z)^\top x\}$.

\section{Conclusion}
We have developed a new learning framework for robust  predictions of  outcomes in a more interpretable  way. Beyond preoperative outcomes, we could use this approach to better predict and understand important postoperative outcomes such as opioid refill~[\cite{sekhri2018probability}], new chronic opioid use~[\cite{brummett2017new}], hospital readmission, and opioid overdose. However, the best way to quantify the uncertainty of the estimates is still unknown.  We will  pursue this later.
\newpage
\renewcommand\thefigure{\thesection\arabic{figure}}    
\setcounter{figure}{0}   
\renewcommand\thetable{\thesection\arabic{table}}    
\setcounter{table}{0}   
\begin{center}
    APPENDIX
\end{center}

\begin{appendix}
\section{Architecture of DNN} 
\begin{figure}[H]
    \centering
    \includegraphics[scale=0.5]{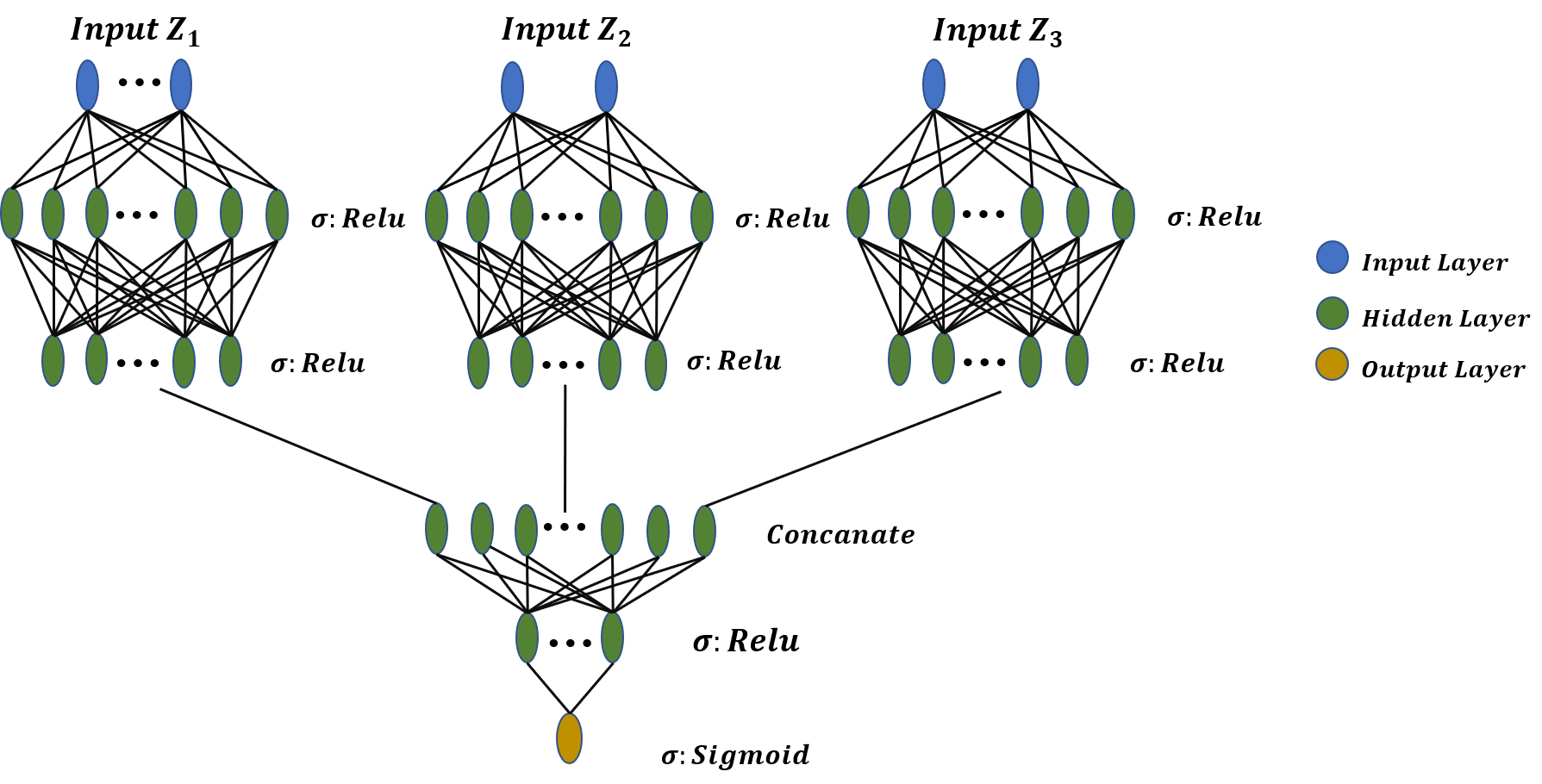}
    \caption{\textbf{Architecture of DNN for the AOS Data.} {\bf Input:}  preoperative characteristics are classified into three inputs: $Z_1$ are un-modifiable characteristics, such as gender and race;  $Z_2$ are modifiable characteristics, such as BMI and smoking;  $Z_3$ are  pain-related characteristics, such as  Fibromyalgia Survey Score and pain severity.
     {\bf Layers:} each category of inputs goes through the same structure: two hidden layers with a ReLu activation function. The first hidden layer has 500 neurons and the the second hidden layer has 125 neurons. The three structures are concatenated and passed onto a layer with 15 hidden neurons and a ReLu activation function. 
     {\bf Output:} estimated probability of preoperative opioid use. 
    }
    \label{Fig4}
\end{figure}
\section{Sensitivity Analysis}
Because stochastic gradient descent is sensitive to the choice of learning rates (LR),
we use grid search to tune the learning rate. For the real data analysis, we tune the learning rate over a range from 0.005 to 0.1 with 20 equally spaced grid points, 
and find that $LR=0.01$ seems to strike a balance between stability and computational readiness. We also implement  the adaptive SGD to analyze our data. Specifically, we have implemented three popular adaptive SGD algorithms, namely,  ``Adagrad'', ``Adadelta'' and ``Adam.''
Adagrad adapts the learning rate based on a sequence of subgradients ~[\cite{duchi2011adaptive}] to improve the robustness of SGD and avoid  tuning  the learning rate manually~[\cite{dean2012large}], while Adadelta~[\cite{zeiler2012adadelta}] and  Adam ~[\cite{kingma2014adam}] only store an exponentially decaying average of  subgradients~[\cite{zeiler2012adadelta}].  We conduct 100 experiments to compare the prediction performance using different optimizers. In each experiment, we randomly split data into the training and testing parts, and use the balanced subsampling strategy described in Section 5.2  to assess the performance on the testing data. The means and standard errors (se)  
of different metrics are summarized in Table \ref{Opt}. We find that all four methods give similar performances, though SGD with a fixed $LR=0.01$ 
and Adam
give  the same C-statistic and sensitivity, slightly better than those obtained  by 
Adagrad and Adadelta; all of these methods give the same balance accuracy.

\begin{figure}[h]
    \centering
    \includegraphics[scale=0.6]{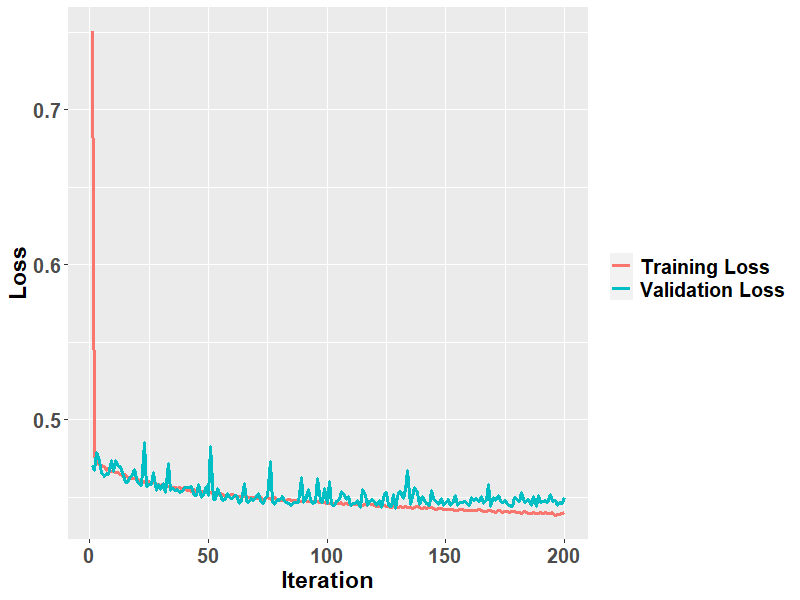}
    \caption{\textbf{Learning Curve of INNER: Cross Entropy Loss Against Iteration For Training and Validation Data}}
    \label{learning_curv}
\end{figure}

\begin{table}[h]
\centering
\addtolength{\leftskip} {-2.5cm}
\addtolength{\rightskip}{-2.5cm}
\begin{threeparttable}
\caption{\label{Opt}
Prediction Performance of INNER Using different Optimizers$^{a,b}$}
\begin{tabular}{lcccc}
\hline
 & \thead{SGD (LR=0.01)} & \thead{Adagrad} & \thead{Adadelta} & \thead{Adam}\\ 
 \hline
C-statistic & 0.78 (0.0006) & 0.77 (0.0005)& 0.76 (0.0006) & 0.78 (0.0006)\\
Accuracy & 0.72 (0.0029) & 0.73 (0.0011) & 0.73 (0.0006) & 0.72 (0.0009)\\
Sensitivity & 0.69 (0.0052) & 0.66 (0.0022) & 0.66 (0.0012) & 0.69 (0.0020)\\
Specificity & 0.73 (0.0052) & 0.76 (0.0019) & 0.76 (0.0010) & 0.73 (0.0017)\\
Balance Accuracy & 0.71 (0.0008) & 0.71 (0.0006)& 0.71 (0.0005) &0.71 (0.0006) \\
\hline
\end{tabular}
\begin{tablenotes}
\item[a.] used the balanced subsampling strategy and a threshold of 0.5
\item[b.] based on   100 random splits
\end{tablenotes}
\end{threeparttable}
\end{table}
The number of iterations is chosen to ensure the convergence of the algorithm (as shown in Fig \ref{learning_curv}).  We have also varied the batch sizes and number of iterations to examine the stability of the
results and find a batch size of 64 and an epoch of 200 give a reasonable performance.
We have conducted sensitivity analysis to assess the robustness of SGD towards the choices of these hyperparameters, and we find that the model's C-statistic is fairly robust to them. Specifically, by varying the learning rate from  0.0075 to 0.0125, the batch size from 32 to 128 and the number of iterations from 200 to 250, the C-statistic of the obtained INNER model is around 0.78.

\begin{table}[h]
\centering
\addtolength{\leftskip} {-2.5cm}
\addtolength{\rightskip}{-2.5cm}
\begin{threeparttable}
\caption{\label{lr}
Average C-statistics (se) of INNER with Various Learning Rates, Batch Sizes and Epochs$^{a,b,c}$}
\begin{tabular}{llccc}
\hline
 && \thead{LR =0.0075 } & \thead{LR = 0.01 } & \thead{LR = 0.0125} \\ 
 \hline
\multirow{3}{*}{BS = 32 } & Epoch = 150  & 0.78 (0.0005) & 0.78 (0.0006) & 0.78 (0.0006) \\
& Epoch = 200 & 0.78 (0.0005) & 0.78 (0.0007) & 0.78 (0.0006) \\
& Epoch = 250 & 0.78 (0.0005) & 0.78 (0.0007) & 0.78 (0.0006) \\
\hline
\multirow{3}{*}{BS = 64} & Epoch = 150 &  0.78 (0.0006) & 0.78 (0.0007) & 0.78 (0.0007)\\
& Epoch = 200 & 0.78 (0.0005) & 0.78 (0.0006) & 0.78 (0.0006) \\
& Epoch = 250 & 0.78 (0.0005) & 0.78 (0.0006) & 0.78 (0.0005) \\
\hline
\multirow{3}{*}{BS =128 } & Epoch = 150 & 0.78 (0.0006) & 0.78 (0.0006) & 0.78 (0.0006) \\
& Epoch = 200 & 0.78 (0.0006) & 0.78 (0.0006)  & 0.78 (0.0006) \\
& Epoch = 250 & 0.78 (0.0006) & 0.78 (0.0005) & 0.78 (0.0006) \\
\hline
\end{tabular}
\begin{tablenotes}
\item[a.]  used the balanced subsampling strategy and a threshold of 0.5
\item[b.] used SGD for optimization
\item[c.]  based on 100 experiments 
\end{tablenotes}
\end{threeparttable}
\end{table}

We have conducted additional sensitivity analyses to examine the performance of the model under various initialization schemes  for the weights $\mathbf{W}$ and the biases $\mathbf{b}$ in the neural networks. We have explored using different weights, such as uniform and normal weights,  for the initial weights ~[\cite{glorot2010understanding,he2015delving}]. In particular, we have studied two versions of uniform weights: for a weight matrix $\mathbf{W}_l \in \mathbf{R}^{k_{l+1} \times k_l}$,
where $k_l$ and $k_{l+1}$ are the numbers of input and output units of the $l$th layer,  we initialize it with $\text{Uniform}\{-\sqrt{6/(k_l + k_{l+1})},\sqrt{6/(k_l + k_{l+1)}}\}$ following \cite{glorot2010understanding}  
(labeled as ``Glorot uniform'' in Table \ref{init}, which reports the sensitivity analysis results); we also initialize the weight matrix with Uniform$(-\sqrt{6/k_l},\sqrt{6/k_l})$ following \cite{he2015delving}
(labeled as ``He uniform'' in Table \ref{init}).   For the normal weights,  
we use Normal$(0, 2/(k_l + k_{l + 1}))$ 
as the initial weights following \cite{glorot2010understanding} (labeled as ``Glorot normal'' in Table \ref{init}). 
Finally, for the bias vector $\mathbf{b}$, we initialize it to be either all 0's or 1's for its components (labeled as ``Zeros'' or ``Ones'' in the column of bias initialization in Table \ref{init}).
 For each set-up, we find that the C-statistic of the model is fairly constant, which is 0.78 with varied initialized values of  weights and biases.


\begin{table}[h]
\centering
\addtolength{\leftskip} {-2.5cm}
\addtolength{\rightskip}{-2.5cm}
\begin{threeparttable}
\caption{\label{init}
Average (se)   C-statistics with  different Initializations of Weights and Biases$^{a,b,c}$}
\begin{tabular}{cccc}
\hline
Weight Initialization & Bias Initialization & \thead{C-statistic}\\ 
 \hline
\multirow{2}{*}{Glorot uniform} & Zeros & 0.78 (0.0006) \\
& Ones & 0.78 (0.0006) \\
\multirow{2}{*}{Glorot normal} & Zeros & 0.78 (0.0005)\\
& Ones & 0.78 (0.0005) \\
\multirow{2}{*}{He uniform} & Zeros & 0.78 (0.0008) \\
    & Ones & 0.78 (0.0006)\\
\hline
\end{tabular}
\begin{tablenotes}
\item[a.] used the balanced subsampling strategy and a threshold of 0.5
\item[b.] used SGD for optimization
\item[c.] based on 100 experiments
\end{tablenotes}
\end{threeparttable}
\end{table}


\begin{table}[H]
\centering
\begin{threeparttable}
\caption{Comparisons of the Prediction Performance using the AOS Data$^{a,b,c}$}
\begin{tabular}{lccc}
\hline
 &\thead{Deep Neural \\Network} &\thead{Logistic \\Regression}&\thead{Interpretable Neural \\Network Regression} \\ 
  \hline
\multicolumn{2}{l}{\textbf{Preoperative Opioid Prevalence: 0.23} }  &  &  \\ 
\quad C-statistic & 0.78 (0.0006) & 0.62 (0.0094) & 0.78 (0.0006) \\ 
\quad \textbf{Threshold = 0.50}&&&\\
  \qquad Accuracy & 0.80 (0.0004) & 0.70 (0.0116) & 0.80 (0.0004) \\ 
  \qquad Sensitivity & 0.33 (0.0049) & 0.43 (0.0331) & 0.31 (0.0057) \\ 
  \qquad Specificity & 0.94 (0.0016) & 0.78 (0.0238) & 0.94 (0.0018) \\ 
  \qquad Balance Accuracy & 0.63 (0.0017) & 0.61 (0.0071) & 0.63 (0.002) \\ 
  \quad \textbf{Threshold = 0.23} &  &  &  \\ 
  \qquad Accuracy & 0.72 (0.0021) & 0.69 (0.0123) & 0.73 (0.0030) \\ 
  \qquad Sensitivity & 0.69 (0.0039) & 0.44 (0.0336) & 0.68 (0.0055) \\ 
  \qquad Specificity & 0.73 (0.0038) & 0.77 (0.025) & 0.74 (0.0054) \\ 
  \qquad Balance Accuracy & 0.71 (0.0006) & 0.61 (0.007) & 0.71 (0.0007) \\ 
  \multicolumn{2}{l}{\textbf{Preoperative Opioid Prevalence: 0.50} }   &  &  \\ 
  \quad C-statistic & 0.78 (0.0006) & 0.73 (0.0027) & 0.78 (0.0006) \\ 
  \quad \textbf{Threshold = 0.50}&&&\\
  \qquad Accuracy &  0.73 (0.0017) & 0.63 (0.0129) & 0.72 (0.0029) \\ 
  \qquad Sensitivity &  0.69 (0.0043) & 0.67 (0.0261) & 0.69 (0.0052) \\ 
  \qquad Specificity &0.73 (0.0034) & 0.62 (0.0238) & 0.73 (0.0052) \\ 
  \qquad Balance Accuracy & 0.71 (0.0007) & 0.64 (0.0049) & 0.71 (0.0008) \\ 
  \quad\textbf{Threshold = 0.23} &  &  &  \\ 
  \qquad Accuracy & 0.46 (0.0044) & 0.50 (0.0130) & 0.41 (0.0056) \\ 
  \qquad Sensitivity & 0.93 (0.0024) & 0.84 (0.0154) & 0.95 (0.0022) \\ 
  \qquad Specificity & 0.31 (0.0064) & 0.39 (0.0211) & 0.24 (0.0080)  \\ 
  \qquad Balance Accuracy &  0.62 (0.0021) & 0.61 (0.0047) & 0.60 (0.0030) \\ 
  \hline
\end{tabular}
\label{AppendixTable1}
\begin{tablenotes}
\item[a.] 
prediction power of each model with the best architectures (DNN and INNER) under different sampling strategies and threshold; 
for the comparison of different architectures, refer to Appendix Table \ref{AppendixTable2} and Appendix Table \ref{AppendixTable3}
\item[b.] based on 100 experiments for each metric
\item[c.] in the AOS data, the prevalence of preoperative opioid is 0.23,  and the prevalence is around 0.23 for the training data; we use the balanced subsampling strategy to adjust the prevalence of preoperative opioid to be 0.50 in the training data
\end{tablenotes}
\end{threeparttable}
\end{table}

\begin{table}[H]
\centering
\begin{threeparttable}
\caption{Tuning the Architecture of INNER with the AOS Data$^{a,b,c,d}$}
\begin{tabular}{lccc}
\hline
 & \thead{Three Layers \\250 Neurons}  & \thead{Four Layers \\500 Neurons} & \thead{Five Layers \\500 Neurons} \\ 
  \hline
\multicolumn{2}{l}{\textbf{Preoperative Opioid Prevalence: 0.23} } &  & \\ 
  \quad C-statistic &0.78 (0.0006) & 0.78 (0.0007) & 0.77 (0.0005)  \\ 
  \quad \textbf{Threshold = 0.50} &&&\\
  \qquad Accuracy &  0.80 (0.0004) & 0.79 (0.0005) & 0.79 (0.0004)\\ 
  \qquad Sensitivity &0.31 (0.0057) & 0.32 (0.0063) & 0.32 (0.0061)\\ 
  \qquad Specificity & 0.94 (0.0018) & 0.94 (0.0021) & 0.93 (0.0019)  \\ 
  \qquad Balance Accuracy & 0.63 (0.0020) & 0.63 (0.0021) & 0.63 (0.0021)  \\ 
  \quad \textbf{Threshold = 0.23} &  &  &  \\ 
  \qquad Accuracy & 0.73 (0.0030) & 0.72 (0.0031) & 0.72 (0.0022)\\ 
  \qquad Sensitivity &  0.68 (0.0055) & 0.68 (0.0054) & 0.68 (0.0043) \\ 
  \qquad Specificity & 0.74 (0.0054) & 0.73 (0.0055) & 0.74 (0.0041) \\ 
  \qquad Balance Accuracy &  0.71 (0.0007) & 0.71 (0.0008) & 0.71 (0.0005) \\ 
  \multicolumn{2}{l}{\textbf{Preoperative Opioid Prevalence: 0.50} } &  & \\ 
  \quad C-statistic & 0.78 (0.0006) & 0.78 (0.0006) & 0.78 (0.0006) \\
  \textbf{ Threshold = 0.50}&&&\\
  \qquad Accuracy & 0.72 (0.0029) & 0.73 (0.0026) & 0.72 (0.0020) \\ 
  \qquad Sensitivity &0.69 (0.0052) & 0.69 (0.0048) & 0.69 (0.0037) \\ 
  \qquad Specificity &0.73 (0.0052) & 0.75 (0.0047) & 0.73 (0.0036)\\ 
  \qquad Balance Accuracy & 0.71 (0.0008) & 0.71 (0.0008) & 0.71 (0.0005) \\ 
  \quad \textbf{Threshold = 0.23} &  &  &  \\ 
  \qquad Accuracy &0.41 (0.0056) & 0.42 (0.0057) & 0.43 (0.0050) \\ 
  \qquad Sensitivity &  0.95 (0.0022) & 0.95 (0.0023) & 0.94 (0.0021)\\ 
  \qquad Specificity &0.24 (0.0080) & 0.26 (0.0081) & 0.28 (0.0071)  \\ 
  \qquad Balance Accuracy &0.60 (0.0030) & 0.60 (0.0030) & 0.61 (0.0026)\\  
  \hline
\end{tabular}
\label{AppendixTable2}
\begin{tablenotes}
\item[a.] 
the first column is for the best INNER architecture as reported in Table \ref{Table2} and Appendix Table \ref{AppendixTable1}
\item[b.] the other columns refer to the other more complicated INNERs, with more hidden layers or more neurons in each hidden layers
\item[c.] the column names are the number of hidden layers and the number of neurons in the first hidden layers for $F_L(\mathbf{Z}_i;\boldsymbol{\alpha})$ and $F_L(\mathbf{Z}_i;\boldsymbol{\alpha})$
\item[d.] in the AOS data, the prevalence of preoperative opioid use is 0.23; 
uses a balanced subsampling strategy by over-sampling cases; adjusts the prevalence of preoperative opioid use to be 0.50 in the training data
\end{tablenotes}
\end{threeparttable}
\end{table}

\begin{table}[H]
\centering
\begin{threeparttable}
\caption{Tuning the Architecture of DNN for AOS Data$^{a,b,c,d}$}
\begin{tabular}{lccc}
\hline
 & \thead{Two Layers\\500 Neurons} & \thead{Three Layer \\ 250 Neurons} & \thead{Three Layer\\500 Neurons} \\ 
  \hline
\multicolumn{2}{l}{\textbf{Preoperative Opioid Prevalence: 0.23} } &  & \\ 
  \quad C-statistic & 0.78 (0.0006) & 0.79 (0.0006) & 0.79 (0.0005)  \\ 
  \quad \textbf{Threshold = 0.50} &&&\\
  \qquad Accuracy & 0.80 (0.0004) & 0.79 (0.0004) & 0.79 (0.0004) \\ 
  \qquad Sensitivity & 0.33 (0.0049) & 0.34 (0.0054) & 0.32 (0.0068) \\ 
  \qquad Specificity & 0.94 (0.0016) & 0.93 (0.0018) & 0.94 (0.0020)  \\ 
  \qquad Balance Accuracy & 0.63 (0.0017) & 0.63 (0.0018) & 0.63 (0.0024) \\ 
  \quad \textbf{Threshold = 0.23} &  &  &  \\ 
  \qquad Accuracy & 0.72 (0.0021) & 0.72 (0.0022) & 0.72 (0.0024) \\ 
  \qquad Sensitivity & 0.69 (0.0039) & 0.70 (0.0038) & 0.69 (0.0049) \\ 
  \qquad Specificity & 0.73 (0.0038) & 0.72 (0.0040) & 0.73 (0.0046) \\ 
  \qquad Balance Accuracy &  0.71 (0.0006) & 0.71 (0.0006) & 0.71 (0.0005) \\ 
  \multicolumn{2}{l}{\textbf{Preoperative Opioid Prevalence: 0.50} } &  & \\ 
  \quad C-statistic & 0.78 (0.0006) & 0.78 (0.0006) & 0.78 (0.0005) \\
  \textbf{ Threshold = 0.50}&&&\\
  \qquad Accuracy & 0.73 (0.0017) & 0.72 (0.0025) & 0.72 (0.0023) \\ 
  \qquad Sensitivity &0.69 (0.0043) & 0.70 (0.0039) & 0.70 (0.0043) \\ 
  \qquad Specificity &0.73 (0.0034) & 0.72 (0.0043) & 0.73 (0.0042) \\ 
  \qquad Balance Accuracy & 0.71 (0.0007) & 0.71 (0.0007) & 0.71 (0.0005) \\ 
  \quad \textbf{Threshold = 0.23} &  &  &  \\ 
  \qquad Accuracy &0.46 (0.0044) & 0.45 (0.0044) & 0.45 (0.0052) \\ 
  \qquad Sensitivity &  0.93 (0.0024) & 0.94 (0.0020) & 0.93 (0.0023) \\ 
  \qquad Specificity &0.31 (0.0064) & 0.30 (0.0062) & 0.31 (0.0074)  \\ 
  \qquad Balance Accuracy & 0.62 (0.0021) & 0.62 (0.0022) & 0.62 (0.0026) \\  
  \hline
\end{tabular}
\label{AppendixTable3}
\begin{tablenotes}
\item[a.] 
the first column is for the best DNN architecture as reported in Table \ref{Table2} and Appendix Table \ref{AppendixTable1} 
\item[b.] the other columns  refer to the other more complicated DNNs, with more hidden layers or more neurons
\item[c.] the column names are the number of hidden layers and the number of neurons in the first hidden layers before concatenation
\item[d.] in the AOS data, the prevalence of preoperative opioid use is 0.23; uses a balanced subsampling strategy by over-sampling cases;  adjusts the prevalence of preoperative opioid use to be 0.50 in the training data
\end{tablenotes}
\end{threeparttable}
\end{table}

\section{Subpopulations with High Risks}
\begin{figure}[H]
    \centering
    \begin{subfigure}[b]{0.3\textwidth}
         \centering
         \includegraphics[width=\textwidth]{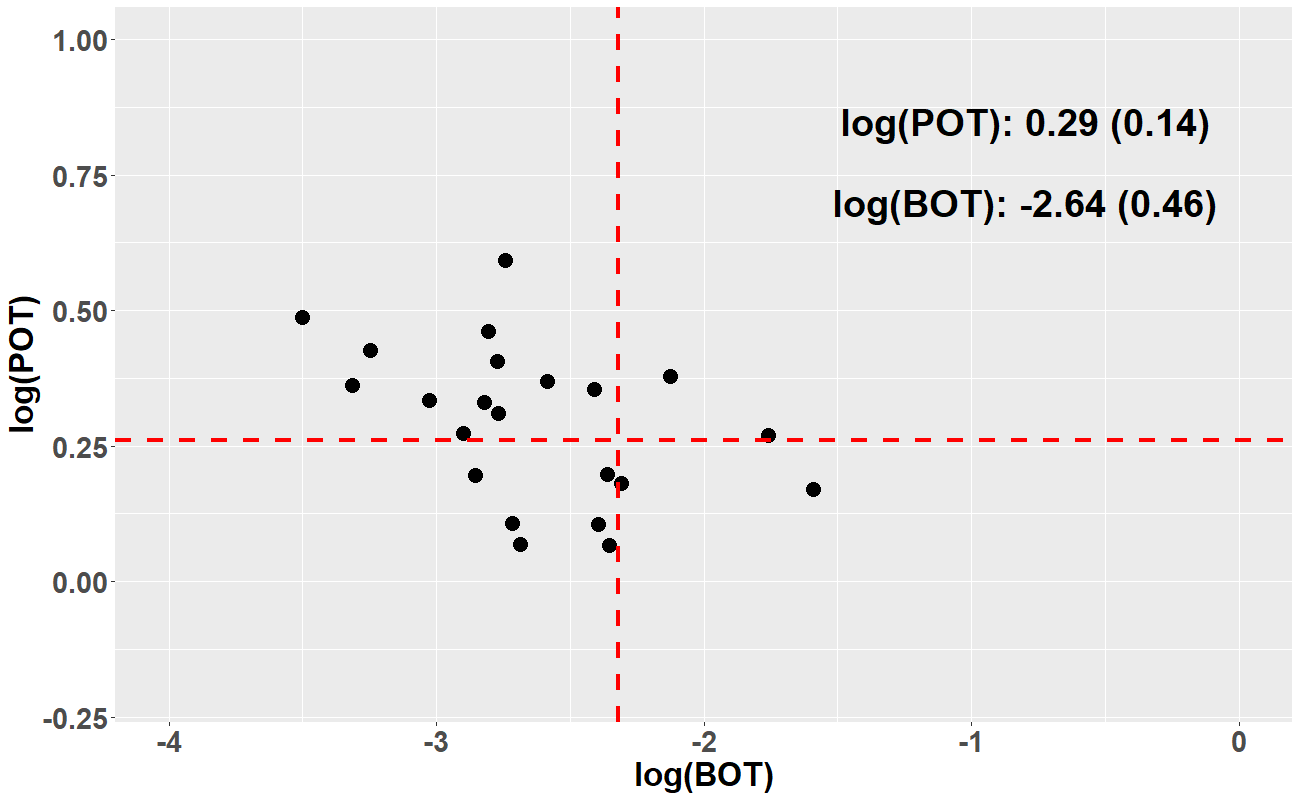}
         \caption{}
         \label{demoSub}
     \end{subfigure}
     \hfill
     \begin{subfigure}[b]{0.3\textwidth}
         \centering
        \includegraphics[width=\textwidth]{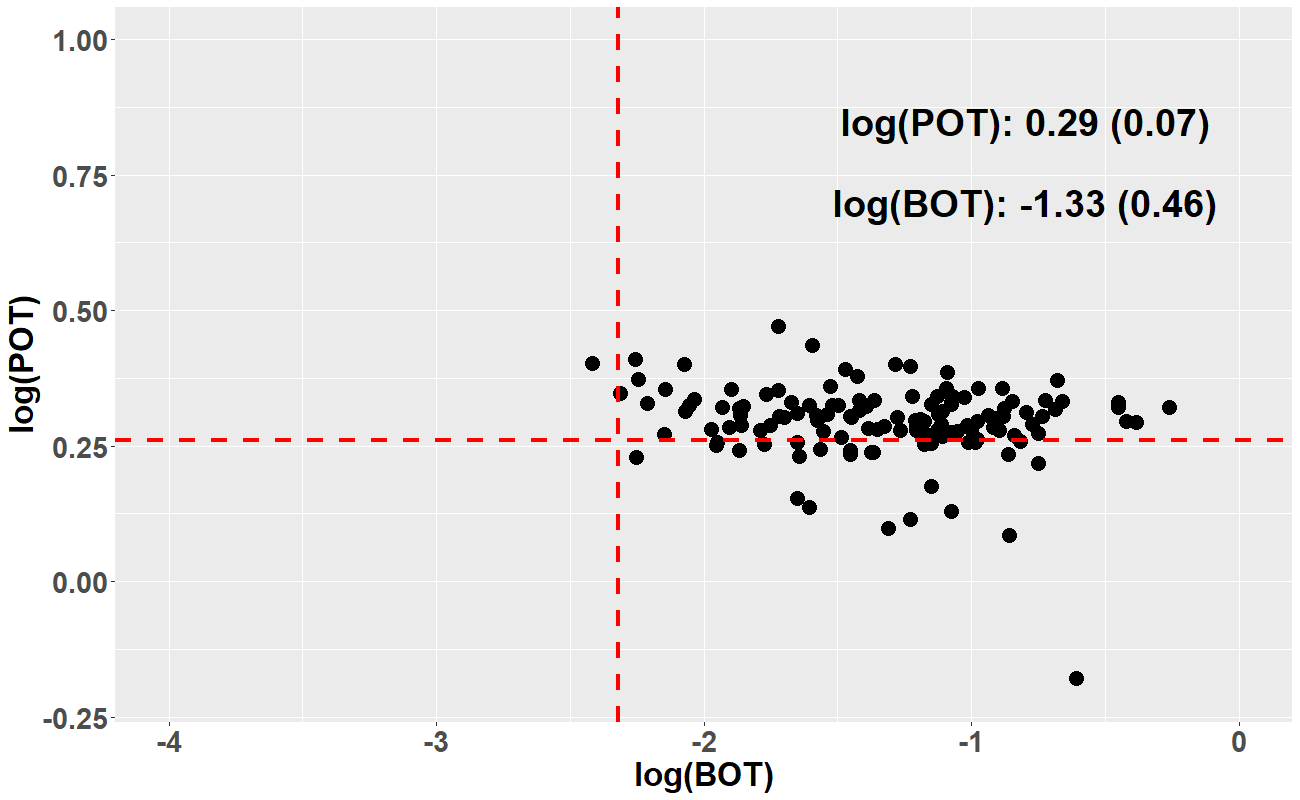}
         \caption{}
         \label{physicalCondSub}
     \end{subfigure}
     \hfill
     \begin{subfigure}[b]{0.3\textwidth}
         \centering
        \includegraphics[width=\textwidth]{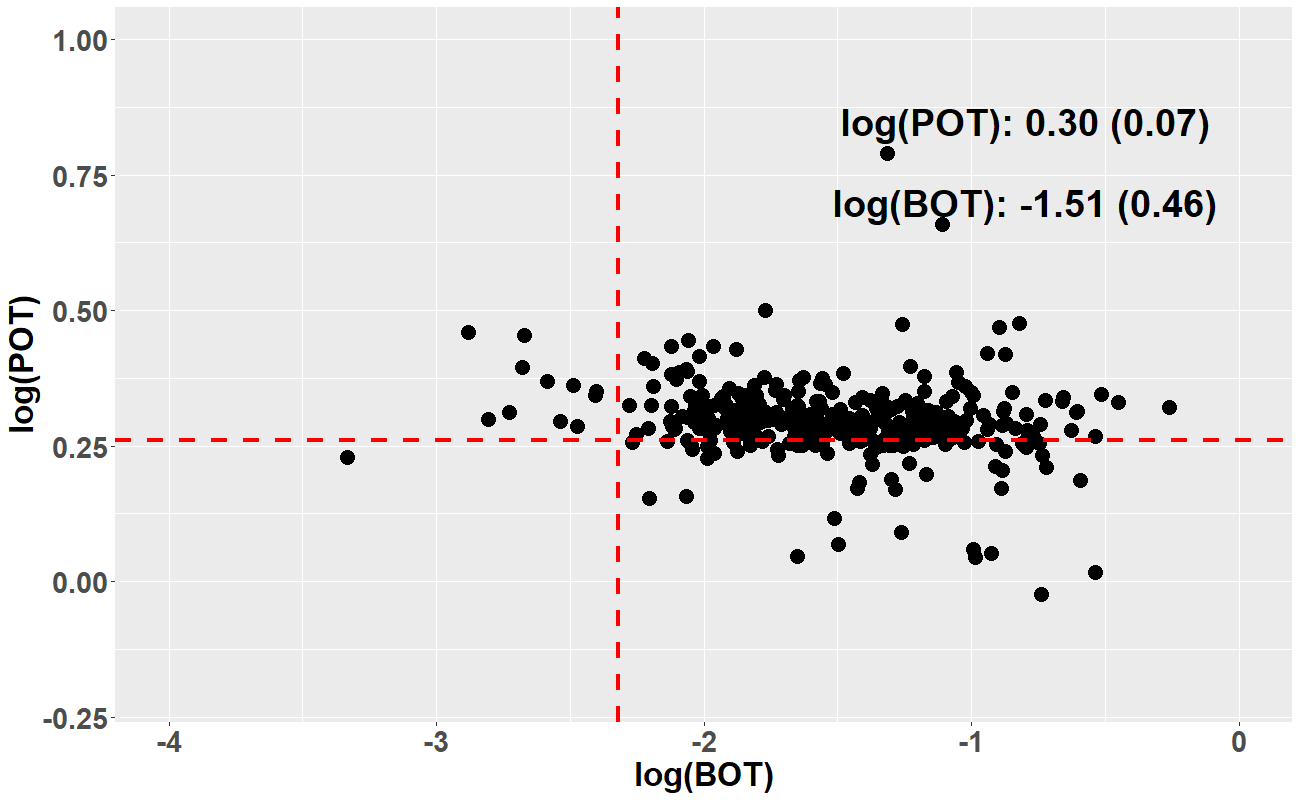}
         \caption{}
         \label{SudSub}
     \end{subfigure}
        \caption{\textbf{Distributions of BOT and POT for three groups.}
        \textbf{(a)}: Patients are chosen based on demographics, including gender, race and age;\textbf{(b)}: Patients are chosen based on physical condition risk factors, including BMI, ASA scores, Fibromyalgia survey scores, Charlson comorbidity index and sleep apnea;\textbf{(c)}: Patients are chosen based on substance use and co-occurring mental disorders, including illicit drug use history, tobacco consumption, depression and anxiety. The horizontal and vertical lines represent the population means of $\log(\text{POT})$ and $\log(\text{BOT})$, respectively; the numbers in each plot refer to the means and standard deviations.
         }
        \label{three groups}
\end{figure}

We have made scatter plots of $\log(\text{POT})$ and $\log(\text{BOT})$ (Fig \ref{three groups}) for the three groups mentioned in Section 5.4. The population means (standard deviation (std)) of $\log(\text{POT})$ and $\log(\text{BOT})$ are 0.26 (0.09) and -2.32 (0.63) respectively.  In  Fig \ref{three groups}(a), we focus on a group of patients identified by  demographic risk factors, i.e.,  African American male patients younger than 20 years old; these patients  on average have a higher $\log(\text{POT})$ (mean: 0.29, std: 0.14) but
a lower $\log(\text{BOT})$ (mean: -2.64, std: 0.46), indicating  their  sensitivity to pain but lower tendency to take opioids without pains. Fig \ref{three groups}(b) depicts BOT and POT for patients who have worsened physical conditions, i.e., with  BMI greater than or equal to 32, ASA scores between three and four, Fibromyalgia survey scores greater than 13, Charlson comorbidity index greater than or equal to one, and sleep apnea;  these patients have a higher $\log(\text{BOT})$ (mean: -1.33, std: 0.46) and higher  $\log(\text{POT})$ (mean: 0.29, std: 0.07) compared to the entire population, indicating they are both sensitive to pain and likely to take preoperative opioids even with no pains reported. Finally, Fig \ref{three groups}(c) focuses on patients who have substance use and co-occurring mental disorders, such as illicit drug use history, tobacco consumption, anxiety and depression. These patients have a smaller $\log(\text{BOT})$ (mean: -1.51, std: 0.46) on average compared to those in Fig \ref{three groups}(b) and the highest $\log(\text{POT})$ (mean: 0.30, std: 0.07) among the three groups.

\end{appendix}

\begin{center}SUPPLEMENTARY MATERIAL \end{center}

\textbf{Code Availability. } Code developed for this project is available at \url{https://github.com/YumingSun/INNER}.

\bibliographystyle{unsrtnat}
\bibliography{references}  






\end{document}